\documentclass[10pt,twocolumn,letterpaper]{article}

\pdfoutput=1

\usepackage{cvpr}
\usepackage{times}
\usepackage{epsfig}
\usepackage{graphicx}
\usepackage{amsmath}
\usepackage{amssymb}
\usepackage{pifont}% http://ctan.org/pkg/pifont

\usepackage{multirow}
\usepackage{eso-pic}

\usepackage{array}
\newcommand{\PreserveBackslash}[1]{\let\temp=\\#1\let\\=\temp}
\newcolumntype{C}[1]{>{\PreserveBackslash\centering}p{#1}}
\newcolumntype{R}[1]{>{\PreserveBackslash\raggedleft}p{#1}}
\newcolumntype{L}[1]{>{\PreserveBackslash\raggedright}p{#1}}

\usepackage[pagebackref=true,breaklinks=true,letterpaper=true,colorlinks,bookmarks=false]{hyperref}

\cvprfinalcopy % *** Uncomment this line for the final submission

\def\cvprPaperID{923} % *** Enter the CVPR Paper ID here

\ifcvprfinal\fi
\begin{document}

%%%%%%%%% TITLE
\title{HVNet: Hybrid Voxel Network for LiDAR Based 3D Object Detection}
% in point cloud
\author{
	\begin{tabular}{ p{2.8cm}<{\centering} p{2.8cm}<{\centering} p{2.8cm}<{\centering}}
Maosheng Ye & Shuangjie Xu & Tongyi Cao \\
\end{tabular}\\
DEEPROUTE.AI\\
{\tt\small \{maoshengye, shuangjiexu, tongyicao\}@deeproute.ai}\\
}

\maketitle
%\thispagestyle{empty}

%%%%%%%%% ABSTRACT
\begin{abstract}
    \vspace{-4pt}
      We present Hybrid Voxel Network (HVNet), a novel one-stage unified network for point cloud based 3D object detection for autonomous driving.
      Recent studies show that 2D voxelization with per voxel PointNet style feature extractor leads to accurate and efficient detector for large 3D scenes. 
      Since the size of the feature map determines the computation and memory cost, the size of the voxel becomes a parameter that is hard to balance.
      A smaller voxel size gives a better performance, especially for small objects, but a longer inference time.
      A larger voxel can cover the same area with a smaller feature map, but fails to capture intricate features and accurate location for smaller objects.
      We present a Hybrid Voxel network that solves this problem by
      fusing voxel feature encoder (VFE) of different scales at point-wise level and project into multiple pseudo-image feature maps.
    %   decoupling the Pointnet feature extraction scale and the pseudo-image projection scale. % wording need 
      We further propose an attentive voxel feature encoding that outperforms plain VFE and a feature fusion pyramid network to aggregate multi-scale information at feature map level.
      Experiments on the KITTI benchmark show that a single HVNet achieves the best \textit{mAP} among all existing methods with a real time inference speed of 31$Hz$.

    \vspace{-10pt}
    \end{abstract}
    
    \begin{figure}
    \vspace{-15pt}
    \centering
    \includegraphics[width=8cm]{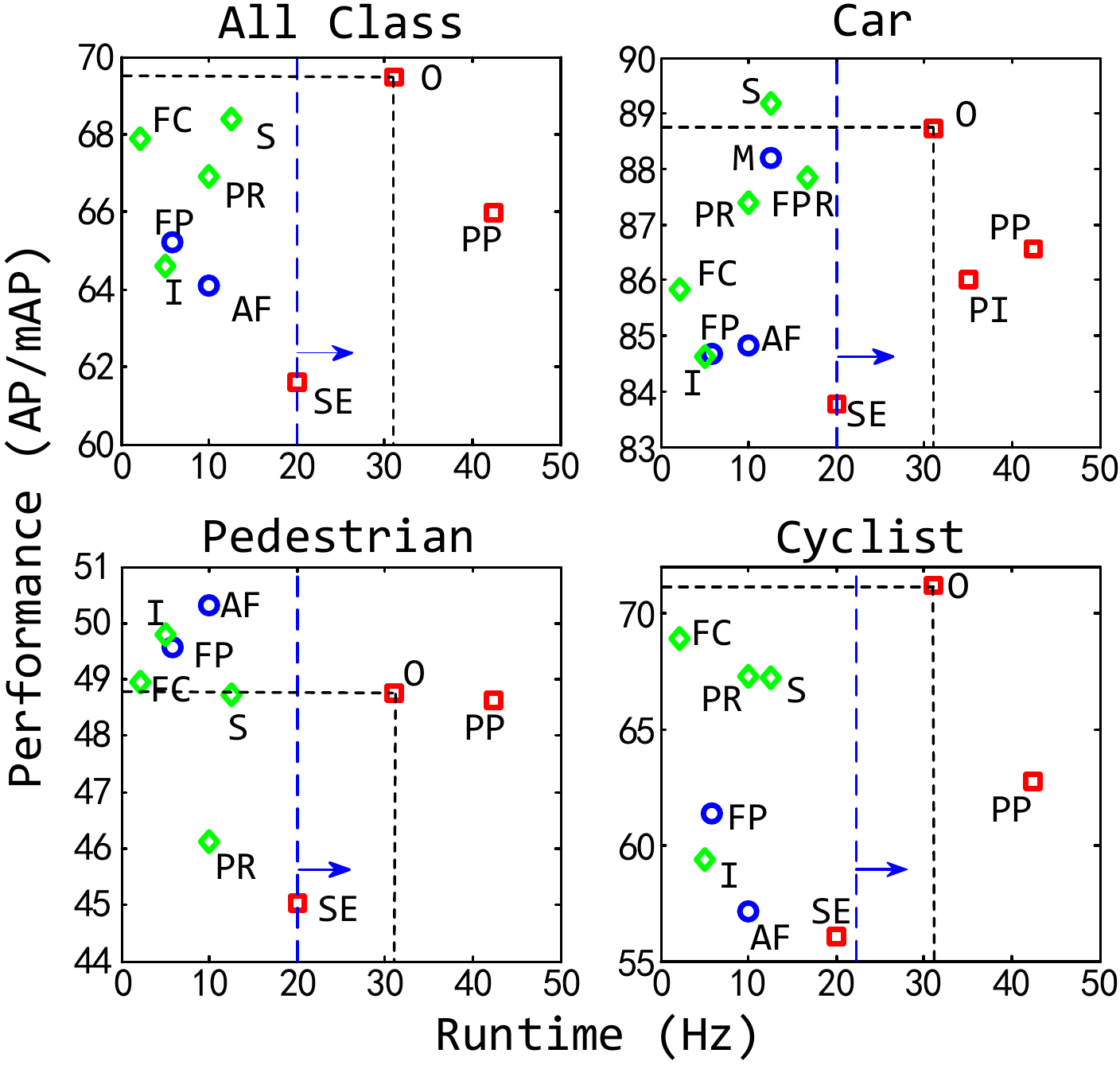}
    \caption{Bird’s eye view performance vs speed in KITTI~\cite{geigerAreWeReady2012} test set. Multi-sensor methods drawn as blue circles; two-stage LiDAR based methods drawn as green diamonds and one-stage methods drawn as red squares. Methods on the right of the blue vertical bar are real-time.
    %Since the KITTI dataset only annotated objects in image view about 90 degree area of the point cloud view, most existing methods which run at 10$Hz$ are far from enough. A reasonable definition for real time should be at least 20$Hz$ for onboard application.
    Methods should be at least 20$Hz$ since onboard application should cover 360 degree rather than KITTI annotation at limited 90 degree.
    Drawn methods are FP: F-PointNet~\cite{qiFrustumPointNets3D2018}, AF: AVOD-FPN~\cite{kuJoint3DProposal2018}, M: MMF~\cite{liangMultiTaskMultiSensorFusion2019}, I: IPOD~\cite{yang2018ipod}, FC: F-ConvNet~\cite{wangFrustumConvNetSliding2019}, S: STD~\cite{yangSTDSparsetoDense3D2019}, PR: PointRCNN~\cite{shiPointRCNN3DObject2019}, FPR: Fast Point R-CNN~\cite{chenFastPointRCNN2019}, SE: SECOND~\cite{yan2018second}, PP: PointPillars~\cite{langPointPillarsFastEncoders2019}, PI: PIXOR++~\cite{yangHDNETExploitingHD2018} and O: our HVNet. For PointPillars we use their runtime on PyTorch for a fair comparison. Our approach outperforms in most case, especially for \textit{mAP} and \textit{Cyclist}.}
    % , especially in the one-stage approaches.
    \label{fig:P-R}
    \vspace{-10pt}
    \end{figure}
    
    %%%%%%%%% BODY TEXT
    \section{Introduction}
    % \xu{scanning density of lidar larger; longer distance ego-car need to see; the inference time is more and more important, with a reasonable accuracy maintained. Thus ...}
    
    % \xu{Voxel-FPN, multi-scale: smaller voxel size bring accuracy but has low speed; inference time increases linearly.}
    
    % \xu{Our main idea: Hybird Voxel Network, encoding multi-scale voxel features, add voxel scale causes little speed loss but a lot performance boost. }
    
    % \xu{Compare: other methods (2d and 3d), use scale under a restriction of scale correspondence; however, point cloud has scaling property natively; Hybird Voxel Net more flexible to take advantage of both the small computation of large scales and the fine representation of small scales, by encoding voxels of hybird scale into scale-invariant global context.}
    
    % Question
    3D object detection is the task to recognize and locate objects in 3D scene. It serves as a fundamental task for 3D scene understanding with wide applications in robotics and self-driving cars.
    Recent approaches utilize various types of data, including  monocular images~\cite{simonelliDisentanglingMonocular3D2019a, xuMultiLevelFusionBased2018}, stereo images~\cite{liStereoRCNNBased2019, qinTriangulationLearningNetwork2019} and point cloud~\cite{langPointPillarsFastEncoders2019, zhouVoxelNetEndtoEndLearning2018} from LiDAR.

    % , among which point cloud is what we focus on due to the common use of LiDAR sensor in autonomous driving.
    % its abundant geometry information.
    % crucial to offer spatial information.
    % Challenge
    Unlike 2D images, point cloud data has some unique properties. 
    % Point cloud provides accurate geometry and robustness to illumination changes, which is essential to 3D localization.
    % Point cloud data provides accurate geometry and depth information.
    %  compared to stereo images. 
    The location of each point reflects the surface of physical objects in the real world. 
    In bird eye view, data is scale invariant inherently and objects are naturally separated, which facilitates the detection of occluded object. 
    Due to its accuracy and robustness, LiDAR has become the prevalent sensor for autonomous.
    %    scale invariance 
    %    In order to utilize the merits of different scales in point cloud. 
    %    Compared to 2D images, the scale or the size of per grid is proportional to the physical dimension of point cloud in the real world. 
    %    Besides that, multi scale in point cloud can be referred to both multi scale feature and multi scale point feature in point cloud. 
    %    Due to the physical meaning, we can also rescale the points in order to zoom the area we want to focus on. 
    %    In order to extract more compact features from highly sparse and irregular LIDAR data, .
    However, the density of point cloud varies greatly. 
    The volume density is proportional to the inverse of square distance to the sensor. 
    This makes it difficult to adopt point cloud segmentation networks or 2D image detection methods. 
    %high sparsity and irregularity due to sensor properties make it difficult to apply regular 2D image detection methods. 
    
    % 3D object detection and localization with point cloud play an important role in the scene understanding for autonomous driving cars. 
    
    % It is more demanding with time efficiency and accuracy. 
    
    % In this work, we propose a novel one-stage unified network to predict object localization and class label with high efficiency compared to existing methods.

    % Current methods and there drawbacks, for our method
    
    % TODO(tongyi): move to ralted work 
    % Some of them~\cite{Chen2016Multi, engelckeVote3DeepFastObject2017, kuJoint3DProposal2018, meyer2019lasernet} convert the point cloud into different views to generate and gather feature representation. 
    
    % For example, LaserNet\cite{meyer2019lasernet} projected the data into sensor range view and apply regular image semantics segmentation techniques for feature extraction. 
    % These approaches generally fail to take full advantage of geometry in the data and suffer from long runtime due to multi-view computation.
    % With emergence of PointNet\cite{qiPointNetDeepHierarchical2017a}, point-based approaches \cite{shiPointRCNN3DObject2019, yangSTDSparsetoDense3D2019} have been popular since they directly operate on point cloud.
    % and don't need any other views conversion which may lead to information loss. 
    % However, memory requirement and computation cost of point-based method is too large for application, and highly related to the number of point.
    
    \begin{figure}
    \vspace{-15pt}
        \centering
        \includegraphics[width=7.6cm]{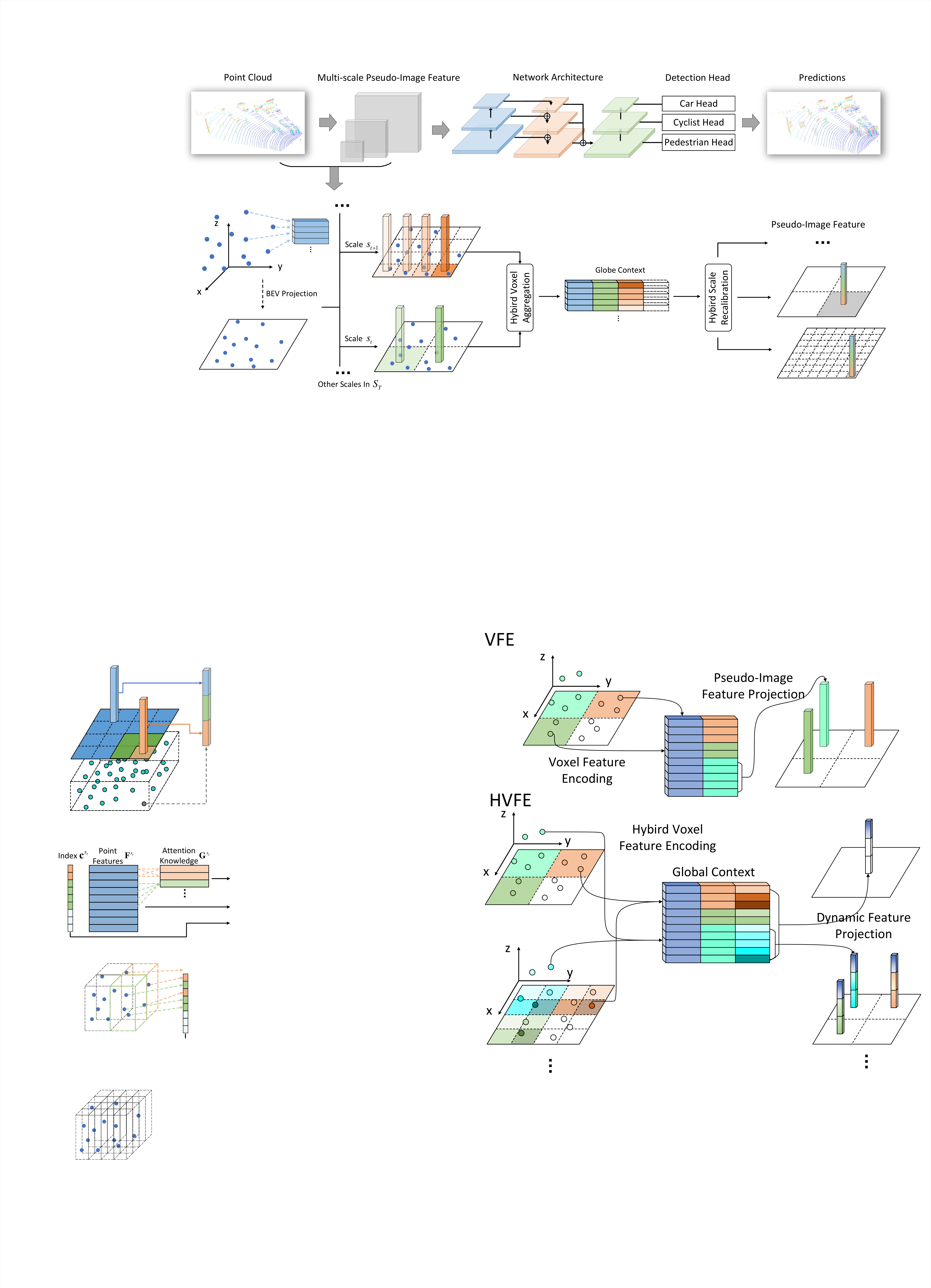}
        \caption{The voxel feature extraction strategy of VFE methods and our HVNet. Each point feature in VFE methods only contains one specified scale knowledge. As to HVNet, point features under hybrid scales are aggregated into a global context, then projected to features in dynamic target scales.}
        \label{fig:multi-scale_2}
    \vspace{-10pt}
    \end{figure}
    In order to handle these problems and utilize the advantages of LiDAR data, many approaches have been proposed recently (see Fig~\ref{fig:P-R} caption). 
    A noticeable line of methods combine 2D/3D voxelization with PointNet feature extractor in each voxel \cite{langPointPillarsFastEncoders2019, yan2018second, zhouVoxelNetEndtoEndLearning2018}, which we call voxel feature encoding (VFE) methods.
    Using learnt point cloud features and 2D pseudo image feature map, these methods achieve both good accuracy and high inference speed.
    VFE methods generally contain three steps (Fig \ref{fig:multi-scale_2} VFE). \textbf{1. Voxelization}: the point cloud is assigned to a 2D voxel grid.
    \textbf{2. Voxel Feature Extraction}: a grid dependent point-wise feature is computed for each point and fed into a Pointnet style feature encoder.
    \textbf{3. Projection}: the point-wise feature is aggregated to the voxel-level feature and projected back to their original grid, forming a pseudo-image feature map. 
    
    A key parameter in VFE methods is the size of voxel. 
    A smaller voxel captures finer geometry feature and better localizes the object but suffers from longer inference time. 
    A coarser voxel leads to a smaller feature map and faster inference speed but has inferior performance, especially for small objects. 
    % For autonomous driving, the objects of interest are car, cyclist and pedestrian, which have the scale varying from 5 meters to 0.5 meters.
    In~\cite{langPointPillarsFastEncoders2019}, a grid size of 0.12 gives the best result on KITTI dataset with a speed of 25 Hz. However, the inference space is limited to the 90 degree FOV in front with a 70 meter range. 
    % The resulting feature map is 600x500. 
    % However for urban driving, a 360 degree perception with range of 120m is standard. This leads to a feature map of size 2000x2000.
    % Doubling the voxel size will reduce the number of operation by 4 times approximately. 
    Other work uses multiple voxel scales to enhance the performance~\cite{wangVoxelFPNMultiscaleVoxel2019} of the network but the runtime is still dominated by the largest feature map.
    % They normally apply pointnet to the points in the same bin to generate pseudo image. After that, a 
    % series of traditional convolution can be used. Indeed, this kind of method gains a fine feature representation for point cloud in a fast way by utilizing more powerful feature extractor rather than hand crafted feature in traditional voxel-based method. However, 
    % However, most of voxel-based methods introduce tremendous loss in both information content and efficiency during the voxelization phase.
    % only a few approaches~\cite{wangVoxelFPNMultiscaleVoxel2019} have focused on the scale invariance, but with a cumbersome method that can only map raw data to the corresponding scale feature one-to-one. In these methods, a fine grid size is often carefully chosen to maintain a decent performance, since the memory cost and runtime cost increases with fine grid size which leads to larger feature map. For example, the feature map size will be 800 x 800, if we want to detect an area of 80m x 80m with voxel size 0.1meter.
    %Indeed, a smaller grid size can capture more details of point cloud in neighborhood. Besides that, the current methods rarely focus on efficiency of real autonomous driving scenes.  
    % generate a birdeye view representation. 
    
    % It is more demanding with time efficiency and accuracy. 
    
    % In this work, we propose a novel one-stage unified network to predict object localization and class label with high efficiency compared to existing methods.

    % Our method, with some comparison, can be multiply paragraph
    In order to utilize fine-grained voxel feature as well as maintain runtime efficiency, we propose Hybrid Voxel Network (HVNet).
    HVNet decouples the feature extraction scale and the pseudo-image projection scale.
    Similar to~\cite{langPointPillarsFastEncoders2019}, the HVNET contains three steps (Fig \ref{fig:multi-scale_2} HVFE): \textbf{1. Multi-Scale Voxelization}: We create a set of feature voxel scale and each point is assigned to multiple voxels. 
    % Note that the voxel is maintained implicitly by storing the voxel indexes rather than by creating the actual dense voxel in memory.
    \textbf{2. Hybrid Voxel Feature Extraction}: for each scale, a voxel dependent feature is computed for each point and fed into the attentive voxel feature encoder (AVFE). Feature from each voxel scale is concatenated point-wise.  \textbf{3. Dynamic Feature Projection} To project the feature back to a pseudo-image, we create another set of multi-scale projection voxels, which can be \emph{independent} of the feature voxels. The multi-scale point-wise feature from the last step is aggregated according to the projection scale voxels, forming multiple pseudo-image feature maps.
    By these strategies, the HVNet is able to be time-efficient.
    As an example, a set of feature voxel scale can be $\{0.1m, 0.2m, 0.4m\}$ and the projection voxel scale can be $\{0.4m\}$. For a $80m\times 80m$ area we end up with a feature map of size $200 \times 200$.
    For a normal VFE structure, we need a feature map of size of $800\times 800$ to capture small feature at $0.1m$ scale.

    We evaluated our HVNet on the KITTI~\cite{geigerAreWeReady2012} dataset. HVNet achieves the state-of-the-art performance among all the LiDAR one-stage methods (Fig.~\ref{fig:P-R}). Besides, even when compared with methods of two stage and multi-sensor, our approach also outperforms in \textit{mAP} and \textit{Cyclist}. At the same time, HVNet runs at a real-time speed of 31$Hz$.
    
    In summary, our contributions reside as follows:
    \begin{itemize}
    \vspace{-4pt}
        \item We propose an novel Hybrid Voxel Feature Encoder that fuses multi-scale VFE efficiently by decoupling the feature extraction voxels and the feature map projection voxels.
    \vspace{-8pt}
        \item We maintain all the voxels implicitly and perform the scale aggregation in the compact point-wise form to achieve high computation and memory efficiency.
    \vspace{-8pt}
        \item We propose an attentive VFE that achieves great improvement with little computation overhead.
    \vspace{-8pt}
        \item Extensive experiments are conducted on the KITTI to demonstrate the effectiveness of each component. 
        % HVNet achieves the state of the art performance. 
    
    \end{itemize}
    
    % \begin{figure*}[t]
    %     % \vspace{-5pt}
    %     \centering
    %     \includegraphics[width=1.0\textwidth]{network}
    %     \caption{Total Pipeline.}
    %     \label{fig:network}
    %     % \vspace{-10pt}
    % \end{figure*}

    \begin{figure*}[t]
    \vspace{-15pt}
        \centering
        \includegraphics[width=0.95\textwidth]{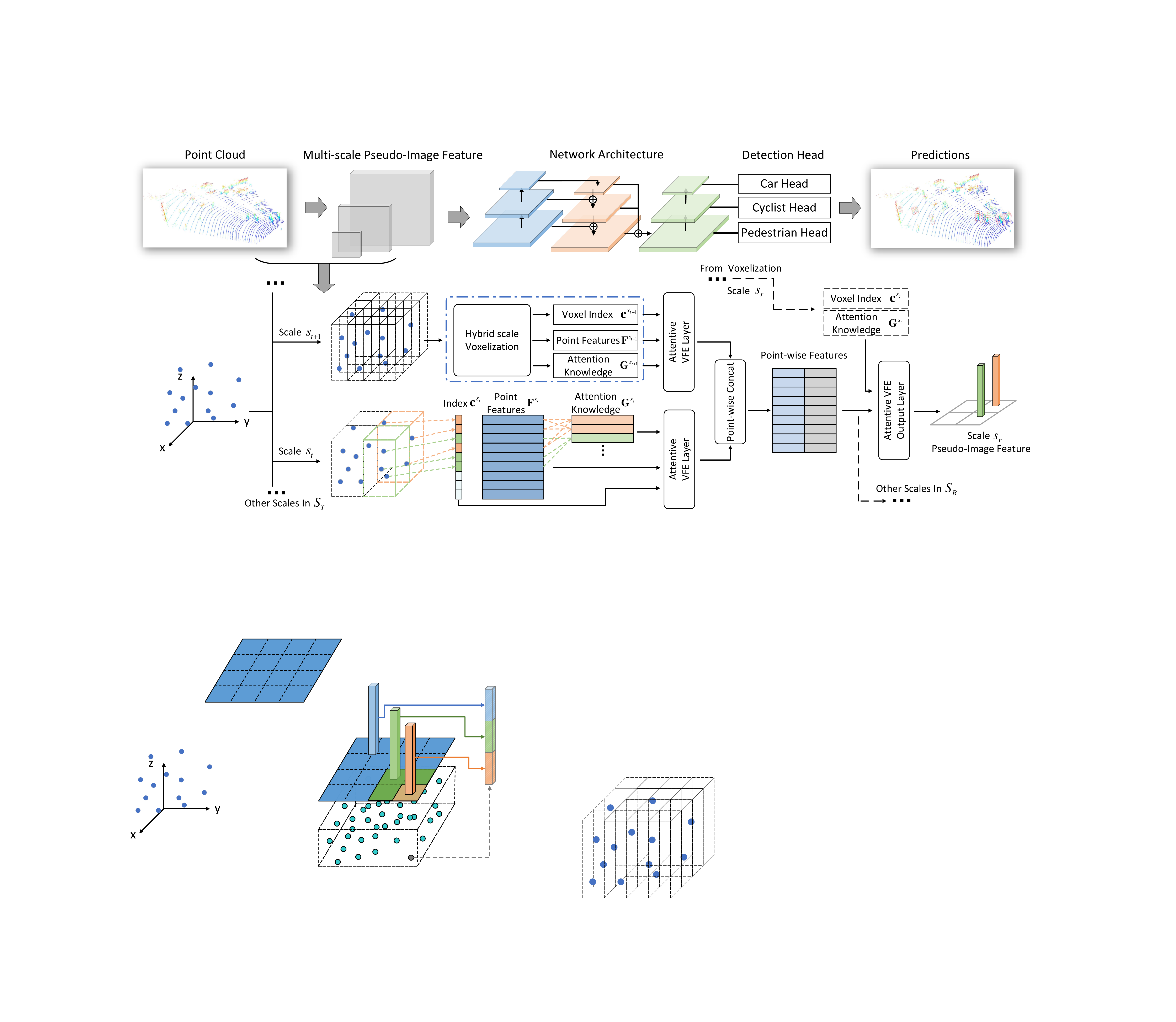}
        \caption{The first line is the whole structure of HVNet. The second line is the Hybrid Voxel Feature Extractor architecture, consisting of four parts: a) multi-scale pre-voxelization in $x$-$y$ plane; b) multi-scale point cloud data encoding by multi-stream AVFE layers parallelly; c) encoded features aggregation; d) dynamic feature projection by AVFEO layers with aggregated features and target scale knowledge.}
        \label{fig:multi-scale}
    \vspace{-10pt}
    \end{figure*}
    
    \vspace{-5pt}
    \section{Related Work}

    \vspace{-3pt}
    \subsection{3D Object Detection}
    \vspace{-5pt}
    % First image-based method: multi-view images (video or stereo data) and monocular image
    
    There are roughly two different lines for existing methods of 3D object detection with point cloud:
    % They are multi-sensor fusion based and point cloud based methods.
    
    % multi-view images (video or stereo data) and monocular image
    
    \textbf{Multi-sensor based 3D object detection.} 
    MV3D \cite{Chen2016Multi} is a pioneering work which projects point cloud into bird eye view and front view.
    %  and then employs a Faster-RCNN~\cite{Ren2015Faster} for feature extraction and 3D object prediction. 
    AVOD~\cite{kuJoint3DProposal2018}, inspired by MV3D, designs 3D object anchors and then aggregates corresponding feature maps by projection to generate a comprehensive representation for detection. 
    F-PointNet~\cite{qiFrustumPointNets3D2018} extends the 2D detections from image into corresponding frustums in the 3D space. 
    MMF~\cite{liangMultiTaskMultiSensorFusion2019} is proposed to exploit multiple related tasks including depth completion and 2D object detection for accurate multi-sensor 3D object detection. 
    However, although multiple sensors could provide extra information, the inference efficiency for these frameworks are relatively low. 
    Besides, synchronization requirement between multiple sensors makes it hard to deploy.
    
    \textbf{Point cloud based 3D object detection.}
    % A major breakthrough~\cite{charlesPointNetDeepLearning2017} in 3D data representation leads to enormous progress in LIDAR-based 3D object detection.
    Voxel based methods~\cite{engelckeVote3DeepFastObject2017, kuJoint3DProposal2018, Yang2018PIXOR, wang2015voting} share a main idea to project sparse point cloud into compact representation.
    VoxelNet~\cite{zhouVoxelNetEndtoEndLearning2018} employs VFE layers based on PointNet for 3D space points characterization. 
    SECOND~\cite{yan2018second} utilizes Sparse Convolution~\cite{Graham20173D} to accelerate VoxelNet and improve performance.
    Based on SECOND, PointPillars~\cite{langPointPillarsFastEncoders2019} applies voxelization only in XY direction to form pseudo-image featurization for point cloud. 
    However, point number in each voxel is limited, leading to information loss. In this work, we propose a novel encoder for lossless 3D representation.
    
    There have been many two-stage works~\cite{yang2018ipod, wangFrustumConvNetSliding2019, yangSTDSparsetoDense3D2019, shiPointRCNN3DObject2019, chenFastPointRCNN2019} recently. 
    Fast Point R-CNN~\cite{chenFastPointRCNN2019} applies two-stage framework exploiting volumetric representation for initial predictions and raw point cloud for refinement. 
    STD~\cite{yangSTDSparsetoDense3D2019} first generates proposals and compact proposal features with PointsPool in the first stage, and then predicts 3D bounding box in the second stage. 
    These methods have high accuracy due to better recall rate, but long inference time (lower than 20$Hz$).
    Therefore, We focus on one-stage unified framework with both time efficiency and accuracy in this paper.
    
    \vspace{-3pt}
    \subsection{Multi-scale feature aggregation}
    \vspace{-5pt}
    
    Multi-scale strategy~\cite{liangDeepContinuousFusion2018a, liangMultiTaskMultiSensorFusion2019, chenFastPointRCNN2019} has been proved to be effective to 3D object detection.
    % PointNet++~\cite{qiPointNetDeepHierarchical2017a} firstly proposes multi-scale grouping to capture multi-scale patterns.
    ContFuse~\cite{liangDeepContinuousFusion2018a} uses continuous convolution to aggregate multi-scale feature maps from different ResNet Block~\cite{he2016deep}. 
    Voxel-FPN~\cite{wangVoxelFPNMultiscaleVoxel2019} employs a RPN-FPN module~\cite{lin2017feature} to aggregate different voxel features generated by voxelization from various voxel sizes. 
    However, the voxel feature maps have strict correspondence to each voxel size, making the multi-scale learning inadequacy and inflexible.
    % amount of calculation grow linearly with the number of scales. 
    Instead, our HVNet, a novel multi-scale zoomable aggregation strategy, is able to zoom scales into a many-to-many relationship.
    %Voxel-FPN, chenFastPointRCNN2019, Contfuse, MMF, Deep Continuous Fusion for Multi-Sensor 3D Object Detection
    
    \section{Approach}
    
    In this paper we describe our one-stage 3D object detector HVNet.
    % , that only accepts point cloud as input to produce very accurate 3D boxes for \textit{Car}, \textit{Cyclist} and \textit{Pedestrian} within one forward propagation.
    %  unlike most existing work which separates these three categories tasks.
    The overall network is presented in the first line of Figure~\ref{fig:multi-scale}, which consists of three main stages: 1) hybrid voxel feature extraction that projects the raw point cloud into pseudo-image feature;
    % Hybrid Voxel Feature Extractor architecture to extract multi-scale pseudo-image feature with graph AVFE layer; 
    2) 2D convolutional middle layers for multi-scale feature fusion; 
    and 3) a detection head that predicts 3D boxes for different classes separately.
    % In the following, we introduce these three main stages in detail.
    % our graph based efficient implementation and 
    
    \vspace{-3pt}
    \subsection{Hybrid Voxel Feature Extractor}
    \label{sec:encoder-decoder}
    \vspace{-5pt}
    
    % With the development of these years, 2D convolutional networks has achieved dramatic progress. 
    % In order to apply 2D convolution in irregular and sparse point cloud, a hybrid voxel feature extractor is proposed to obtain a more compact pseudo-image features.
    The three main steps included in the hybrid voxel feature extractor (HVFE), i.e., hybrid scale voxelization, hybrid voxel feature encoding and dynamic feature projection.
    Furthermore, to decouple the hybrid voxel scale and pseudo-image feature projection scale for a more flexible balance between speed and accuracy, the global context for hybrid voxel representation is introduced under the voxel-wise attention guidance, whose detail will be described in in Sec.~\ref{sec:avfe}.
    Overall structure of this stage is shown in Fig.~\ref{fig:multi-scale}.
    
    \textbf{Point cloud representation.} 
    A point cloud would be represented by an unordered point set $\left\{ {{{\bf{p}}_1},{{\bf{p}}_2}, \ldots ,{{\bf{p}}_N}} \right\}$ with ${{\bf{p}}_i} \in {\mathbb{R}^d}$, which includes ${{\bf{x}}_i} \in {\mathbb{R}^3}$ for point coordinate $(x, y, z)$ and ${{\bf{f}}_i} \in {\mathbb{R}^{d-3}}$ for associated feature vector. 
    % ${\overrightarrow f _i}$ is $(r)$ in generally,  where $r$ represents the reflectivity. 
    The physical dimension of the scene that we detect objects on is defined as $L \times W \times H$, from the minimum $\left( {{x_{min }},{y_{min }},{z_{min }}} \right)$ to the maximum $\left( {{x_{max }},{y_{max }},{z_{max }}} \right)$.
    
    % \textbf{3D data augmentation.} 
    
    \textbf{Hybrid scales voxelization.} 
    HSV is a way of voxelization that assigns points to evenly spaced grid of voxels.
    The assignment phase in prior methods, like VoxelNet~\cite{zhouVoxelNetEndtoEndLearning2018} and PointPillar~\cite{langPointPillarsFastEncoders2019}, is accompanied by buffer allocation with a fixed size. Points will be dropped when the buffer capacity for a voxel is exceeded, causing randomness and information loss.
    Instead, we maintain only the voxel index of each point.
    Assume that the point cloud is discretized into numerous voxels ${\bf{V}} = \left\{ {{{\bf{v}}_j}|j = 1, \ldots ,{N_{\bf{v}}}} \right\}$ with resolution of ${v_L} \times {v_W} \times {v_H}$ per voxel. 
    ${v_H}$ is set to ${{H}}$, followed by~\cite{langPointPillarsFastEncoders2019}.
    The point-wise index ${\bf{c}}_i$ for target voxel scale size $s$ is recorded as a cursor to which voxel it lies in, given as:
    \begin{equation}
    \setlength{\abovedisplayskip}{1pt}
    \setlength{\belowdisplayskip}{1pt}
        {{\bf{c}}_i^{\left( s \right)}} = \left\lfloor {\frac{{\left( {x_i - {x_{min }}} \right)}}{{{v_L}s}}} \right\rfloor \left\lfloor {\frac{W}{{{v_W}s}}} \right\rfloor {\rm{ + }}\left\lfloor {\frac{{\left( {y_i - {y_{min}}} \right)}}{{{v_W}s}}} \right\rfloor.
    \end{equation}
    Only the point to voxel mapping relation is needed in our approach, unlike MVF~\cite{zhou2019end} which need bi-directional relationships. In this way, only tensor $\bf{c}$ are allocated during voxelization stage. Besides, we keep all points retained inside a voxel other than methods, which also becomes the precondition for voxel attention.
    
    % \begin{figure}
    % % \vspace{-5pt}
    % \centering
    % \includegraphics[width=8.3cm]{motivation}
    % \caption{Voxel scale study on BEV of KITTI validation set.}
    % \label{fig:voxel_scale}
    % \vspace{-10pt}
    % \end{figure}

    \textbf{Hybrid voxel feature extraction.}
    The HVFE module provides a novel way to create fine-grained point cloud feature with little computation cost.
    Take PointPillar~\cite{langPointPillarsFastEncoders2019} as an example (Fig~\ref{fig:multi-scale_2}), previous voxel based methods are limited to the fixed voxel scale $s$, where each encoded point feature only contains the scale knowledge of $s$.
    Besides, the size of final projected features is the same as the corresponding voxel size, which means that the computation complexity will rise rapidly with the addition of fine-grained voxel scale.
    Assume that the projected pseudo-image feature ${{\bf{I}}^{s}}\in{\mathbb{R}^{{N_W} \times {N_L} \times {N_H}}}$, where ${N_L} = \left\lfloor {{L \mathord{\left/ {\vphantom {L {\left( {{v_L}s} \right)}}} \right.
    \kern-\nulldelimiterspace} {\left( {{v_L}{s}} \right)}}} \right\rfloor$, ${N_W} = \left\lfloor {{W \mathord{\left/ {\vphantom {W {\left( {{v_W}r} \right)}}} \right. \kern-\nulldelimiterspace} {\left( {{v_W}{s}} \right)}}} \right\rfloor$ and $N_H$ is the projection output dimension. When $s$ reduces $n$ times, $N_L \times N_W$ will increase $n^2$ times.
    
    %  takes hybrid voxel scales ${S_T} = \left\{ {{s_t}|1 < t < {N_T}} \right\}$
    
    Shown in Fig~\ref{fig:multi-scale}, the HVFE module takes point features of a set of voxel scales ${S_T} = \left\{ {{s_t}|1 < t < {N_T}} \right\}$ as input, encoding them as unified point-wise feature ${\bf{H}}$ with information from different voxel scales.
    % This process resembles the `zoom' of point cloud features from scale $S_T$ to $S_R$.
    Drawn in Fig.~\ref{fig:multi-scale}, firstly the raw point cloud is hybrid voxelized to get cursor set ${{\bf{c}}^{S_T}} = \{ {{\bf{c}}_i^{\left( {{s_t}} \right)}|{s_t} \in {S_T},i \in \left[ {1,N} \right]} \}$
    % ${C_T} = \left\{ {{C_{{t_i}}}|1 < i < {N_T}} \right\}$ for various scales $S_T$
    . 
    Secondly, attention feature set ${\bf{G}}^{S_T}$ whose tensor dimensionality is same as the pointnet features ${\bf{F}} \in {\mathbb{R}^{N \times  q }}$
    %  of ${\bf{g}}_i^{\left( {{s_t}} \right)}$
    % ${G_T}$ with ${G_{t_i}}\in{\mathbb{R}^{N \times 2d}}$ 
    are obtained. 
    Thirdly, the AVFE layer takes $\bf{F}$, ${\bf{G}}^{S_T}$ and ${\bf{c}}^{S_T}$ together to encode raw data into voxel scale feature set ${\bf{H}}^{S_T}$ specified to each voxel scale $s$ with ${{\bf{H}}^{s_t}} \in {\mathbb{R}^{N \times 2q}}$. 
    All the AVFE layers for different scales share the same parameter. 
    After that, ${\bf{H}} \in {\mathbb{R}^{N \times e}}$ is aggregated by concatenation of the ${\bf{H}}^{S_T}$, where $e=2q{N_T}$. 
    It's validated by experiments that only limited $N_T$ of input multi-scale features is able to bring enough information to retrieve features of unseen scales out of $S_T$, utilizing the dynamic feature projection.
    % H = \left\{ {{h_i}|1 < i < N} \right\}
    
    \textbf{Dynamic feature projection.}
    The global context $\bf{H}$ is a high level feature containing information of hybrid voxel scales.
    Therefore, the feature projection scales ${S_R} = \left\{ {{s_r}|1 < r < {N_R}} \right\}$ of pseudo-image features that are out of $S_T$ can be projected dynamically and flexibly with $\bf{H}$. 
    In order to restore information at scales $s_r$ from ${\bf{H}}$, some scale-related information needs to be used as guidance in the projection, containing point-wise index ${\bf{c}}^{s_r}$
    %  of ${\bf{c}}_i^{{{s_r}}}$
    % ${C_R} = \left\{ {{C_{{r_i}}}|1 < i < {N_R}} \right\}$
     and the attention knowledge 
    %  $G_R$ with ${G_{r_i}}\in{\mathbb{R}^{N \times 2d}}$
     ${\bf{G}}^{s_r}$
    %   of ${\bf{g}}_i^{{{s_r}}}$
    . 
    % For a target scale $r\in{S_R}$, 
    The attentive VFE output (AVFEO) layer takes ${\bf{H}}$, ${\bf{c}}^{s_r}$ and ${\bf{G}}^{s_r}$ to retrieve the voxel-wise feature and then project it to a pseudo-image feature ${{\bf{I}}^{s_r}}$.
    Same as the AVFE layer, all AVFEO layers share the same parameters.
    In the dynamic feature projection, ${\bf{G}}^{s_r}$ is the key for the scale-wise feature retrieval, guiding the AVFEO layer to be attentive to project ${\bf{H}}$ into corresponding scale $s_r$.
    Besides, ${\bf{c}}^{s_r}$ is used for the efficient graph-based implementation of the AVFEO layer (Sec.~\ref{sec:avfe}).
    % Details for the AVFEO layer are in .

    \begin{figure}
    \vspace{-15pt}
    \centering
    \includegraphics[width=8cm]{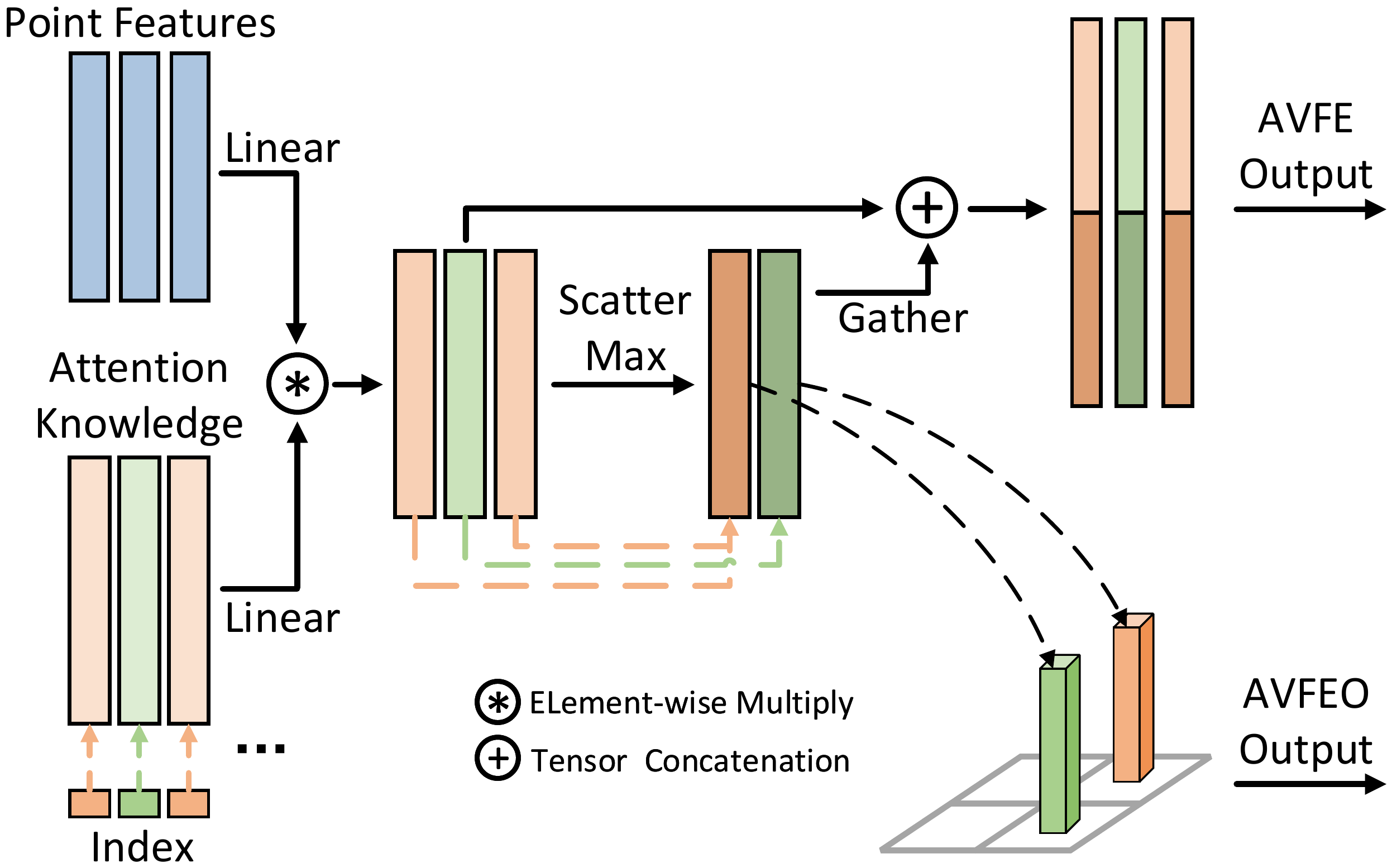}
    \caption{
    AVFE and AVFEO layers. The index tensor is used in indexing operation: Gather and Scatter. 
    % Input point features and gathered attention knowledge are first transformed to the same dimension, and then multiplied together. After that, the voxel-wise feature is obtained by Scatter Max. The obtained feature will be projected to a pseudo-image if AVFEO layer or concatenated with the multiplied point features if AVFE layer.
    }
    \label{fig:vfe}
    \vspace{-10pt}
    \end{figure}
    
    \vspace{-3pt}
    \subsection{Attentive Voxel Feature Encoding}
    \label{sec:avfe}
    \vspace{-5pt}

    \textbf{Standard VFE layer.}
    Traditional VFE~\cite{zhouVoxelNetEndtoEndLearning2018,langPointPillarsFastEncoders2019} first voxelizes data with point number in each voxel limited and allocates fixed preassigned tensors for each voxel. 
    After that, linear transforms and max pooling are conducted in each voxel to aggregate features.
    % , similar to PointNet's architecture~\cite{charlesPointNetDeepLearning2017}. 
    Then pseudo-image features are collected according to the physical location of voxels.
    
    While VFE method provides a simple means of integrating sparse point cloud, it has several drawbacks:
    \begin{itemize}
        \vspace{-5pt}
        \item There are points dropped stochastically during voxelization, leading to tremendous information losses and instability prediction behaviours;
        \vspace{-8pt}
        \item Zero padding is conducted when point number is lower than limitation, which can greatly enlarge amounts of computation and memory consumption;
    \end{itemize}
    
    \textbf{Voxel-wise attention feature.}
    As the expansion of raw point-wise features ${\bf{p}}_i$
    % ${\bf{F}} \in {\mathbb{R}^{N \times d}}$ with $N$ features ${{\bf{f}}_i} \in {\mathbb{R}^d}$
    , attention knowledge ${\bf{g}}_j \in {\mathbb{R}^{q}}$ for voxel ${\bf{v}}_j$ is obtained by voxel-wise feature aggregation, which guides our feature encoding network to concern more about the interrelationships within a voxel. Given index ${\bf{c}}$, the attention knowledge ${\bf{g}}_i$ of point ${\bf{p}}_i$ is represented as:
    % $F = \left\{ {{f_i}|i = 1, \ldots ,N} \right\}$
    % $C{\rm{ = }}\left\{ {{c_i}|1 < i < N} \right\}$
    \begin{equation}
    \setlength{\abovedisplayskip}{1pt}
    \setlength{\belowdisplayskip}{1pt}
        {{\bf{g}}_i} = \left( {\left( {{{{\bf{x}}}_i} - \frac{{\sum\nolimits_{{U_j}} {{{{{\bf{x}}} }_k}} }}{{{N_{{U_j}}}}}} \right) \oplus {{{\bf{f}}}_i}} \right) \oplus \frac{{\sum\nolimits_{{U_j}} {{{\bf{p}}_k}} }}{{{N_{{U_j}}}}},
        % {g_i} = \left( {\left( {{{\overrightarrow x }_i} - \frac{1}{{{N_{{U_i}}}}}\sum\nolimits_{{U_i}} {{{\overrightarrow x }_j}} } \right) \oplus {{\overrightarrow f }_i}} \right) \oplus \frac{1}{{{N_{{U_i}}}}}\sum\nolimits_{{U_i}} {{f_j}} 
    \label{equ:attention_mean}
    \end{equation}
    where $\oplus$ represents the tensor concatenate. The collection $U_j$, referred to the index set of points that lies in the same voxel as ${\bf{p}}_i$, is denoted as $U_j = \left\{ {k|0 < k < N \& {{\bf{c}}_k} = {{\bf{c}}_i}} \right\}$
    % \begin{equation}
        % U_j = \left\{ {k|0 < k < N \& {{\bf{c}}_k} = {{\bf{c}}_i}} \right\}
    % \label{equ:voxel_set}.
    % \end{equation}
    . The collection process of $U$ has a lot of loops, making this aggregation step unbearably slow. We will discuss the efficient index-based implementation in Sec.~\ref{sec:graph_implementation}.
    
    Shown in the Eq.~\ref{equ:attention_mean}, the attention knowledge for a point $\bf{p}$ in a voxel $\bf{v}$ is composed of two parts: a) the mean of features from all the points lie in $\bf{v}$; and b) the feature of $\bf{p}$ whose point coordinate ${\bf{x}}$ is subtracted by the average coordinate of $\bf{v}$. The attention feature $\bf{g}$ participates in both encoder and decoder as the bridge information.
    %  together with voxel-wise index $c$ between encoder-decoder.
    
    \textbf{Learning to be flexible and attentive.}
    % Recently, researchers have begun to use lossless voxelization methods. 
    % MVF~\cite{zhou2019end} proposes a dynamic voxelization strategy utilizing bi-directional relationships between every pair of ${\bf{v}}_j$ and ${\bf{p}}_i$. 
    % However, MVF will suffer the redundant information when points in voxel is overmuch. 
    % Instead, we propose the pre-voxelization strategy named `pre-voxelization' (PV) to establish the point-wise cursor ${\bf{c}}_i$, which lays a natural foundation for the attentive voxel feature aggregation.
    Shown in Fig.~\ref{fig:vfe}, 
    % the attention knowledge $\bf{G}$ is gathered from $\mathbb{R}^{N_{v} \times 2d}$ to $\mathbb{R}^{N \times 2d}$ according to the cursor $C$. 
    the input point features ${\bf{F}}$ and the attention knowledge ${\bf{G}}$ are conducted by linear layers, followed by a element-wise matrix multiply. 
    After that, multiplied features $\bf{M}$ in the same voxel are aggregated by max pooling layer to form the voxel-wise features ${\bf{G}}^{'} \in \mathbb{R}^{N \times q}$. 
    For AVFE layer, the output 
    % ${\bf{H}} \in \mathbb{R}^{N \times \left( {q + d} \right)}$ 
    is the point-wise concatenation of $\bf{M}$ and ${\bf{G}}^{'}$. 
    For AVFEO layer, each feature of ${\bf{G}}^{'}$ is projected to a pixel $\left( {{x_I^{{{{s_r}}}}},{y_I^{{{{s_r}}}}}} \right)$ in pseudo image ${\bf{I}}^{s_r}$ 
    % in line with the cursor $C$
    , given as:
    % $x_I^{{s_r}} = {\bf{c}}_j^{{s_r}}\mid \left\lfloor {{W \mathord{\left/ {\vphantom {W {{v_W}{s_r}}}} \right. \kern-\nulldelimiterspace} {{v_W}{s_r}}}} \right\rfloor$ and $y_I^{{s_r}} = {\bf{c}}_j^{{s_r}}{\mkern 1mu} {\kern 1pt} {\kern 1pt} \,\bmod \,{\mkern 1mu} {\kern 1pt} {\kern 1pt} \left\lfloor {{W \mathord{\left/ {\vphantom {W {{v_W}{s_r}}}} \right. \kern-\nulldelimiterspace} {{v_W}{s_r}}}} \right\rfloor$
    \begin{equation}
    \setlength{\abovedisplayskip}{1pt}
    \setlength{\belowdisplayskip}{1pt}
        x_I^{{s_r}} = {\bf{c}}_j^{{s_r}}\mid \left\lfloor {\frac{W}{{{v_W}{s_r}}}} \right\rfloor ,y_I^{{s_r}} = {\bf{c}}_j^{{s_r}}\,\bmod \,\left\lfloor {\frac{W}{{{v_W}{s_r}}}} \right\rfloor, 
    \end{equation}
    where $s_r$ is the target scale, $\mid$ denotes exact division and $\bmod$ denotes the module operation.
    
    % Thanks to the HSV operation, HVNet is able to handle inconsistent voxels flexibly instead of fixed-capacity voxels, preventing from the suffering of stochastic data abandon. Besides, entire points are referred to the final embedding with the guidance from attention knowledge, keeping features of voxel with excessive points from redundancy.
    
    \vspace{-3pt}
    \subsection{Efficient Index-based Implementation}
    \label{sec:graph_implementation}
    \vspace{-5pt}
    
    The Hybrid Voxel Feature Extraction module is a index-based learning system, where the irregular graph data (point cloud) is grouped by physical correlation. 
    In order to handle the sparse structures, we propose HSV operation to transform sparse matrix into dense matrix and corresponding indexes.
    Verified by experiments, the efficiency of the HVNet then hinges heavily on how well The index strategy of $U$ and the parallel stream processing schedules. 
    Therefore, the key index-based propagation operators, Scatter and Gather, is implemented on GPUs. 
    
    The Gather conducts the sparse point data propagation within voxel and behave as tensor slice according to the cursor vector $\bf{c}$. For implementation, tensor slice operation in PyTorch ~\cite{paszke2017automatic} is fast enough for Gather.
    
    The Scatter manipulates all values from source tensor into output at the indices specified in $\bf{c}$ along a given axis. In our approach, Scatter Mean is used in voxel wise attention and Scatter Max is used in AVFE and AVFEO layers. Take Scatter Max as an example: ${\rm{ou}}{{\rm{t}}_i} = \max \left( {{\rm{ou}}{{\rm{t}}_i},\max \left( {\left\{ {{\rm{sr}}{{\rm{c}}_j}{\rm{|}}{{\bf{c}}_j} = i,j \in \bf{c}} \right\}} \right)} \right)$
    % \begin{equation}
    %     {\rm{ou}}{{\rm{t}}_i} = \max \left( {{\rm{ou}}{{\rm{t}}_i},\max \left( {\left\{ {{\rm{sr}}{{\rm{c}}_j}{\rm{|}}{{\bf{c}}_j} = i,j \in \bf{c}} \right\}} \right)} \right)
    % \end{equation}
    , where $\rm{out}$ and $\rm{src}$ are output and input respectively, $\bf{c}$ is the `group-index' that references the location of $\rm{src}$. In the implementation of Scatter, atomic lock of GPU Global Memory is used to ensure the consistency of $\arg \max$ results for reproducibility.
    
    \vspace{-3pt}
    \subsection{Backbone Network}
    \vspace{-5pt}
    
    \begin{figure}
    \vspace{-15pt}
    \centering
    \includegraphics[width=7cm]{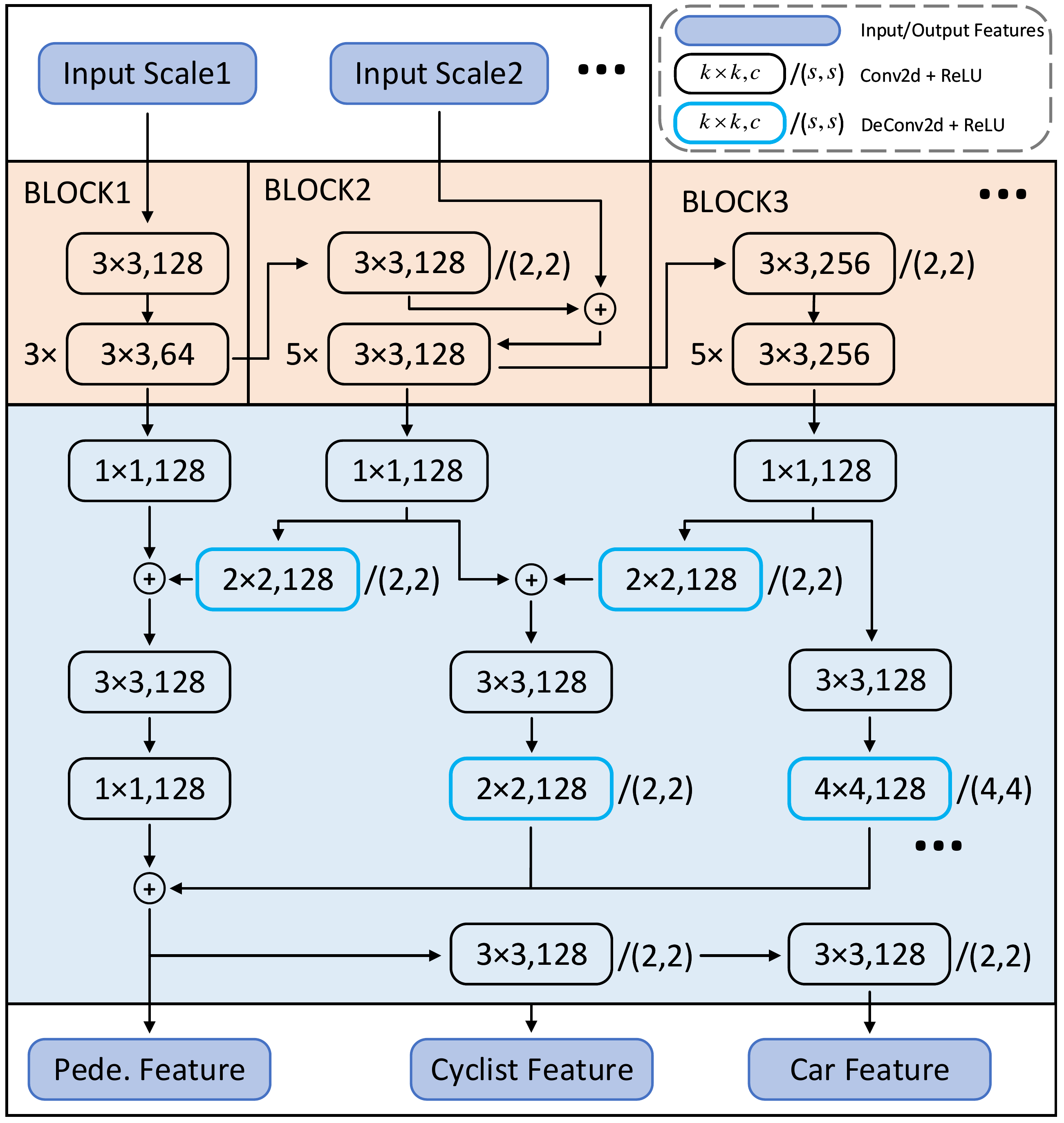}
    \caption{The backbone network architecture of HVNet. Multi-scale features are first fused shallowly in the main stream network, and then fused deeply in the proposed FFPN network.}
    \label{fig:backbone}
    \vspace{-10pt}
    \end{figure}
    
    We use a 2D convolutional backbone network for further detection.
    % and the overall structure is shown in Fig.~\ref{fig:backbone}. 
    The backbone network has two main parts: 1) the main stream network that aggregates multi-scale features ${\bf{I}}^{S_R}$; 2) the FFPN network that refines the feature map and generates multi-class features in the same time.
    
    \textbf{Main stream network with multi-scale feature aggregation
    .} The main stream is shown in the orange part in Fig.~\ref{fig:backbone}, characterized by a concatenation of several blocks. 
    % $\text{Block}\_\{1,2,3\}$. 
    % Each block is composed by several 2D convolutional layers with $3\times3$ kernel size, each followed by a ReLU activation. 
    In $\text{Block}\_2$ and $\text{Block}\_3$, the first layer has a $2\times2$ stride to reduce spatial resolution of features. 
    Given pseudo-image feature set ${\bf{I}}^{S_R}$ where 
    % ${S_R} = \left\{ {r_1},\dots,{r_n} \right\}$ and ${r_{i+1}} = 2{r_{i}}$ 
    ${s_{r + 1}} = 2{s_r}$, ${\bf{I}}^{s_1}$ is taken as input of $\text{Block}\_1$ and ${\bf{I}}^{s_2}$ is aggregated in $\text{Block}\_2$ by tensor concatenation with output of the first layer in $\text{Block}\_2$. 
    More scales can be added by the similar way.
    We take the output of the last three blocks ${{{\bf{B}}_1} \in {\mathbb{R}^{{N_W} \times {N_L} \times {C_1}}}}$, ${{{\bf{B}}_2} \in {\mathbb{R}^{\frac{{{N_W}}}{2} \times \frac{{{N_L}}}{2} \times {C_2}}}}$ and ${{{\bf{B}}_3} \in {\mathbb{R}^{\frac{{{N_W}}}{4} \times \frac{{{N_L}}}{4} \times {C_3}}}}$ as the input of FFPN.
    
    \begin{table*}
    \vspace{-15pt}
    \small
    \centering
    \begin{tabular}{c | c |c c |c  c  c | c  c  c | c  c  c}
    \hline
    \multirow{2}*{\bf Method} & {\bf Speed } & \multicolumn{2}{c|}{\bf Input} & \multicolumn{3}{c|}{\bf \textit{Car} BEV AP (\%)} & \multicolumn{3}{c|}{\bf \textit{Pedestrian} BEV AP (\%)} & \multicolumn{3}{c}{\bf \textit{Cyclist} BEV AP (\%)} \\ \cline{3-13}
        & (Hz) & {L} & {I} & {\textit{Easy}} & {\textit{Moder.}} & {\textit{Hard}} & {\textit{Easy}} & {\textit{Moder.}} & {\textit{Hard}} & {\textit{Easy}} & {\textit{Moder.}} & {\textit{Hard}} \\  
        \hline
    % AVOD~\cite{kuJoint3DProposal2018}  & 12.5 & \checkmark & \checkmark & 89.75   & 84.95   & 78.32   & 42.58   & 33.57   & 30.14   & 64.11   & 48.15   & 42.37  \\
    F-PointNet~\cite{qiFrustumPointNets3D2018}  & 5.9 & \checkmark & \checkmark & 91.17   & 84.67   & 74.77   & 57.13   & 49.57   & 45.48   & 77.26   & 61.37   & 53.78  \\
    AVOD-FPN~\cite{kuJoint3DProposal2018}  & 10 & \checkmark & \checkmark & 90.99   & 84.82   & 79.62   & 58.49   & {\bf 50.32}   & {\bf 46.98}   & 69.39   & 57.12   & 51.09  \\
    MMF~\cite{liangMultiTaskMultiSensorFusion2019} & 12.5 & \checkmark & \checkmark & 93.67   & 88.21   & 81.99  & N/A  & N/A  & N/A & N/A & N/A  & N/A \\
        \hline
        \hline
    IPOD~\cite{yang2018ipod} & 5 & \checkmark &  & 89.64   & 84.62   & 79.96   & {\bf 60.88}   & 49.79   & 45.43   & 78.19   & 59.40   & 51.38  \\
    F-ConvNet~\cite{wangFrustumConvNetSliding2019} & 2.1 & \checkmark &  & 91.51   & 85.84   & 76.11   & 57.04   & 48.96   & 44.33   & {\bf 84.16}   & 68.88   & 60.05  \\
    STD~\cite{yangSTDSparsetoDense3D2019} & 12.5 & \checkmark &  & {\bf 94.74}   & {\bf 89.19}   & {\bf 86.42}   & 60.02   & 48.72   & 44.55   & 81.36   & 67.23   & 59.35  \\
    PointRCNN~\cite{shiPointRCNN3DObject2019} & 10 & \checkmark &  & 92.13   & 87.39   & 82.72   & 54.77 & 46.13 & 42.84 & 82.56 & 67.24 & 60.28 \\
    Fast Point R-CNN~\cite{chenFastPointRCNN2019} & 16.7 & \checkmark &  & 90.87   & 87.84   & 80.52   & N/A & N/A & N/A & N/A & N/A & N/A \\
        \hline
        \hline
    SECOND~\cite{yan2018second} & 20 & \checkmark &  & 89.39   & 83.77   & 78.59   & 55.99   & 45.02   & 40.93   & 76.50   & 56.05   & 49.45  \\
    HRI-VoxelFPN~\cite{wangVoxelFPNMultiscaleVoxel2019} & 50 & \checkmark &  & 92.75   & 87.21   & 79.82   & N/A   & N/A & N/A &N/A  & N/A & N/A \\
    PointPillars~\cite{langPointPillarsFastEncoders2019} & 42.4 & \checkmark &  & 90.07   & 86.56   & 82.81   & \color{blue}57.60   &48.64   & 45.78   & 79.90   & 62.73   & 55.58  \\
    PIXOR++~\cite{yangHDNETExploitingHD2018} & 35 & \checkmark &  & \color{blue}93.28   & 86.01   & 80.11  & N/A & N/A& N/A & N/A & N/A &N/A \\
    \hline
    Ours & 31 & \checkmark &  & 92.83   & \color{blue}88.82   & \color{blue}83.38   & 54.84   &  \color{blue}48.86   & \color{blue}46.33   & \color{blue}83.97   & {\bf 71.17}   & {\bf 63.65}  \\
    \hline
    \end{tabular}
    \caption{Performance of bird eye views on KITTI test set. Here `L' denotes LiDAR input and `I' denotes RGB image.
    We compare with detectors on the KITTI leaderboard evaluated by 40 recall positions. 
    Methods are divided into three types: LiDAR \& image, two-stage LiDAR only and one-stage. The bold results means the best in all methods and the blue results means the best among one-stage methods.} %\protect\footnotemark[1]
    \label{tab:test_results}
    \vspace{-12pt}
    \end{table*}
        %\footnotetext[1]{On 08.10.2019, KITTI team have followed the suggestions of the paper and use 40 recall positions instead of the 11 recall positions proposed in the original Pascal VOC benchmark.}
    
    \textbf{Feature fusion pyramid network.}
    The FPN~\cite{lin2017feature} has been proved to be a valid means for multi-scale feature embedding. Due to the sparse data distribution in point cloud and the small resolution of $I_R$, the FPN structure plays a more important role in our approach. Therefore, an enhanced feature fusion pyramid network is proposed, whose main structure is shown in Fig.~\ref{fig:multi-scale} Instead fusing features from top
    to bottom layer by layer in \cite{lin2017feature}, we first concatenate features at the smallest scale to obtain an intermediate feature ${\bf{B}}_f$, represented as:
    \begin{equation}
    \setlength{\abovedisplayskip}{1pt}
    \setlength{\belowdisplayskip}{1pt}
        {{\bf{B}}_f}{\rm{ = }}\sum\limits_{i = 1}^{{N_B}} {{\Psi _i}\left( {{{\bf{B}}_i} + {1_{i < {N_B}}}{\Gamma _{i + 1}}\left( {{{\bf{B}}_{i + 1}}} \right)} \right)},
        \label{equ:rpn1}
    \end{equation}
    where $+$ means tensor concatenate, $1$ means the indicator function, $N_B$ is the number of ${\bf{B}}_i$, $\Gamma _{i}$ represents the deconv function of each input feature map ${\bf{B}}_i$ for scale alignment and $\Psi _i$ denotes the conv function before being concatenated together. The class-specific pyramid features are given as ${\bf{B}}_o^{\left( i \right)} = {\Upsilon _i}\left( {{{\bf{B}}_f}} \right)$
    % \begin{equation}
    %     {\bf{B}}_o^{\left( i \right)} = {\Upsilon _i}\left( {{{\bf{B}}_f}} \right)
    %     \label{equ:rpn2}
    % \end{equation}
    , where $\Upsilon _i$ denotes the conv layers with various stride. 
    % The pyramid features are obtained in a similar way as 
    Compared with F-SSD~\cite{li2017fssd}, we fuse features in two stages: a) layer by layer fusion in Eq.~\ref{equ:rpn1} and b) downsample convolutional chain. 
    % in Eq.~\ref{equ:rpn2}. 
    Furthermore, within one forward propagation, class-specific pyramid features are able to be obtained, where ${\bf{B}}_o^{\left( 1 \right)}$ for \textit{Pedestrian} class, ${\bf{B}}_o^{\left( 2 \right)}$ for \textit{Cyclist} class and ${\bf{B}}_o^{\left( 3 \right)}$ for \textit{Car} class respectively.

    \vspace{-3pt}
    \subsection{Detection Head and Loss Design}
    \vspace{-5pt}
    
    We use the detection head in the SSD~\cite{liu2016ssd} to detect 3D objects as~\cite{langPointPillarsFastEncoders2019}. In our setting, the positive anchors are selected by matching to the ground truth by Rotated Intersection over Union (RIoU)~\cite{everingham2010pascal, zhouIoULoss2D2019} in bird eye view (BEV). 
    % The localization of bounding box except the elevation participates in the matching process. 
    Each pyramid feature ${\bf{B}}_o^{\left( i \right)}$ is wraped by three parallel branches which are $3 \times 3$ convolution layers $\phi _c^{\left( i \right)}$, $\phi _l^{\left( i \right)}$ and $\phi _h^{\left( i \right)}$ to get classification probability, location offsets and height regression respectively, whose output channel numbers are $N_{\text{anc}} \times N_{\text{cls}}$, $N_{\text{anc}} \times 8$ for location corner offsets in BEV and $N_{\text{anc}} \times 2$ for z center and height. 
    
    Different from most voxel-based methods~\cite{langPointPillarsFastEncoders2019,zhouVoxelNetEndtoEndLearning2018,wangVoxelFPNMultiscaleVoxel2019} that predict center $x,y$ and $w,h,l,\theta$, HVNet utilizes location corner offsets relative to anchors in BEV as localization objective, represented as $\Delta \overrightarrow {\mathrm{cor}}  = \left\{ {\Delta {{\overrightarrow x }_i},\Delta {{\overrightarrow y }_i}|i = 1,2,3,4} \right\}$.
    % \begin{equation}
    %     \Delta \overrightarrow {\mathrm{cor}}  = \left\{ {\Delta {{\overrightarrow x }_i},\Delta {{\overrightarrow y }_i}|i = 1,2,3,4} \right\}
    % \end{equation}
    % , instead of center coordinates and $w,h,l,\theta$. 
    $\Delta \mathrm{cor}$ is a vector of $\mathbb{R}^{8}$ during propagation. Suppose that the location branch $\phi _l$ predicts the offset $\Delta \mathrm{cor}$, then the localization loss is given by ${{\cal L}_{{\rm{loc}}}} = \sum {{\rm{SmoothL1}}\left( {\Delta \overrightarrow {{\rm{cor}}} {\rm{ - }}\Delta {\rm{cor}}} \right)}$.
    % \begin{equation}
    %     {{\cal L}_{{\rm{loc}}}} = \sum {{\rm{SmoothL1}}\left( {\Delta \overrightarrow {{\rm{cor}}} {\rm{ - }}\Delta {\rm{cor}}} \right)}.
    % \end{equation}
    
    As to the classification branch $\phi _c$, given the class probability $p_a$ of an anchor, focal loss~\cite{linFocalLossDense2017a} is used to handle the unbalance between positive and negative samples, represented as ${{\cal L}_{cls}} =  - {\alpha }{\left( {1 - {p_a}} \right)^\gamma }\log \left( {{p_a}} \right)$.
    % The elevation and height regression is separated from location.
    Given prediction ${z,h}$ from $\phi _h$, the vertical loss is denoted as ${{\cal L}_h} = \sum_{b \in \left( {z,h} \right)} {{\rm{SmoothL1}}\left( {\Delta b} \right)}$. 
    
    Therefore, the total loss is given by 
    % ${\cal L} = \frac{1}{{{N_{{\rm{pos}}}}}}\left( {{\lambda _{{\rm{loc}}}}{{\cal L}_{{\rm{loc}}}} + {\lambda _{{\rm{cls}}}}{{\cal L}_{{\rm{cls}}}} + {\lambda _h}{{\cal L}_h}} \right)$.
    
    \begin{equation}
    \setlength{\abovedisplayskip}{1pt}
    \setlength{\belowdisplayskip}{1pt}
        {\cal L} = \frac{1}{{{N_{{\rm{pos}}}}}}\left( {{\lambda _{{\rm{loc}}}}{{\cal L}_{{\rm{loc}}}} + {\lambda _{{\rm{cls}}}}{{\cal L}_{{\rm{cls}}}} + {\lambda _h}{{\cal L}_h}} \right).
    \end{equation}
    
    \begin{figure*}
    \vspace{-15pt}
    \centering
    \includegraphics[width=0.95\textwidth]{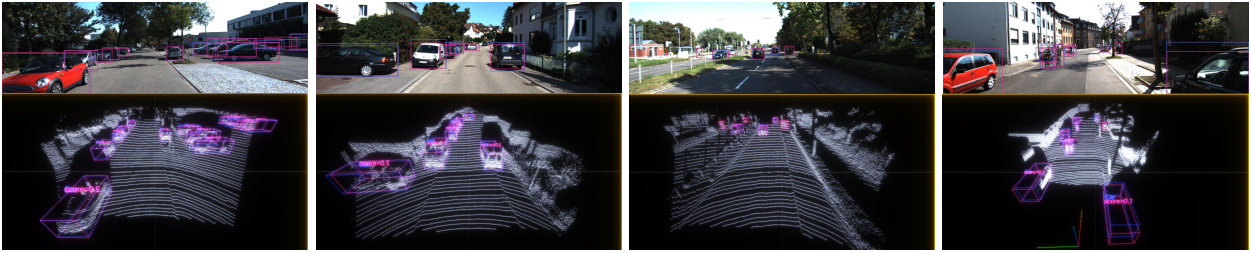}
    \caption{Qualitative analysis on KITTI validation set with Kitti Viewer~\cite{yan2018second}. We show 3D bounding boxes on point cloud along with projected 2D bounding boxes on image. In each image, blue boxes indicates the ground truth, red boxes indicates detections by HVNet.}
    % Qualitative analysis on KITTI. We show a bird’s eye view of the LIDAR point cloud (top), as well as the 3D bounding boxes projected into the image for clearer visualization. Note that our method only uses LIDAR. We show ground truth (gray) and predicted boxes for car (orange), cyclist (red) and pedestrian (blue). The box orientation is shown by a line from the bottom center to the front of the box.
    \label{fig:qualitative}
    \vspace{-5pt}
    \end{figure*}
    
    \vspace{-6pt}
    \section{Experiments}
    \vspace{-3pt}
    
    HVNet is evaluated on the challenging KITTI benchmark~\cite{geigerAreWeReady2012}. We first introduce the setup details of HVNet in Sec.~\ref{sec:setup}. In Sec.~\ref{sec:results}, we compare HVNet with state-of-the-art methods. A detailed ablation study is also provided in Sec.~\ref{sec:ablation} to verify the validity of each component. 
    % In the last subsection Sec.~\ref{sec:visualization}, various visualizations of HVNet and its results are given to bring a intuitive sense.
    
    \vspace{-3pt}
    \subsection{Setup}
    \label{sec:setup}
    \vspace{-5pt}
    
    \textbf{Dataset.}
    KITTI dataset~\cite{geigerAreWeReady2012} consists of 7481 training images and 7518 test images as well as the corresponding point cloud with categories \textit{Car}, \textit{Pedestrian} and \textit{Cyclist}. 
    
    \textbf{Metric.}
    KITTI's metric is defined as average precision (AP) over the 40 recall positions on the PR curve~\cite{everingham2010pascal}. 
    % However, in the test split, 40 recall positions~\cite{simonelli2019disentangling} is adopted by the KITTI team. 
    Labels are divided into three subsets (\textit{Easy}, \textit{Moderate}, \textit{Hard}) on the basis of object size, occlusion and truncation levels. 
    The leaderboard rank is based on results of \textit{Moderate} difficult.
    
    \textbf{Experiment details.}
    The physical detection range is limited within minimum 
    % $l_s=0$, $w_s=-32$, $h_s=-3$
    $\left( {0, - 32, - 3} \right)$ and maximum 
    % $l_m=64$, $w_m=32$, $h_m=2$
    $\left( {64, 32, 2} \right)$. 
    The size of a voxel is $v_L=0.2$, $v_W=0.2$, thus the resolution of BEV feature is $320\times320$. 
    In the encoder and decoder, scale set $S_T=\{0.5, 1, 2\}$ and $S_R=\{1,2,4\}$. Besides, feature dimension is $q=64$ for ${\bf{H}}^{s_t}$ and $N_H=128$ for ${\bf{I}}^{s_t}$. 
    Anchor size is designed as $[0.8,0.8,1.7]$ for \textit{Pedestrian}, $0.8,1.8,1.5$ for \textit{Cyclist} and $[1.7, 3.5, 1.56]$, $[2.0, 6.0, 1.56]$ for \textit{Car}. 
    Each class has the same anchor orientation in $[0, {\pi  \mathord{\left/{\vphantom {\pi  4}} \right. \kern-\nulldelimiterspace} 4}, {\pi  \mathord{\left/{\vphantom {\pi  2}} \right.\kern-\nulldelimiterspace} 2}, 3{\pi  \mathord{\left/{\vphantom {\pi  4}} \right.\kern-\nulldelimiterspace} 4}]$. 
    In the training phase, we choose anchors that have RIoU with ground truth larger than $[0.35, 0.35, 0.5]$ for \textit{Pedestrian}, \textit{Cyclist} and \textit{Car} respectively as positive samples, and those lower than $[0.25, 0.25, 0.35]$ as negative samples. 
    As to the test phase, prediction score threshold is set to 0.2, and the rotated NMS~\cite{neubeck2006efficient} threshold is set to $[0.02, 0.02, 0.4]$.
    In the loss design, $\alpha$ and $\gamma$ for focal loss are set to $[0.75, 0.75, 0.25]$ and $[2., 2., 2.]$ respectively. 
    Three loss weights are $\lambda _{{\rm{loc}}}=1$, $\lambda _{{\rm{cls}}}=1$ and $\lambda _{{\rm{h}}}=1.5$.
     HVNet is trained for 70 epochs with Adam~\cite{kingma2014adam} optimizer, the initial learning rate is ${\rm{lr}}=2{e^{ - 4}}$ with weight decay $1{e^{ - 4}}$. 
     We first employ warmup strategy \cite{he2016deep} with $300$ warmup iterations and ${{1.} \mathord{\left/ {\vphantom {{1.} 3}} \right. \kern-\nulldelimiterspace} 3}$ warmup ratio. 
     Besides, $\rm{lr}$ decays in a ratio of $0.1$ in the $40$-th and $60$-th epoch.
    
    \textbf{3D data augmentation.}
    Global random flipping, rotation, scaling and translation are first applied to the whole point cloud data, where the flipping probability is set to 0.5, the rotation angle ranges are uniform distribution from $[ - {\pi  \mathord{\left/ {\vphantom {\pi  2}} \right. \kern-\nulldelimiterspace} 2}, {\pi  \mathord{\left/ {\vphantom {\pi  2}} \right. \kern-\nulldelimiterspace} 2}]$, the scaling ratio is between $[0.95, 1.05]$ and the location translation obeys normal distribution with mean $0$ and standard deviation $[0.2, 0.2, 0.2]$ for $(x,y,z)$. 
    Followed by SECOND~\cite{yan2018second}, several new boxes from ground truth and corresponding points in other frames, where $8$ for \textit{Cyclist}, $8$ for \textit{Pedestrian} and $15$ for \textit{Car}, will be fixed into current training frame except boxes which have physical collision with boxes in the current frame.
    %. After that, new added boxes overlaped with boxes in current frame will be removed.
    
    \vspace{-3pt}
    \subsection{Experimental Results}
    \label{sec:results}
    \vspace{-5pt}
    
    \begin{table}
    \vspace{-8pt}
    \small
    \centering
    \begin{tabular}{C{2.0cm}|C{0.6cm} C{0.6cm} C{0.6cm}|C{0.6cm} C{0.6cm} C{0.6cm}}
    \hline
    \multirow{2}*{\bf Method} & \multicolumn{3}{c|}{\bf \textit{Car} BEV AP(\%)} & \multicolumn{3}{c}{\bf \textit{Car} 3D AP(\%)} \\
     \cline{2-7} 
     & {\textit{Easy}} & {\textit{Moder.}} & {\textit{Hard}} & {\textit{Easy}} & {\textit{Moder.}} & {\textit{Hard}} \\
     \hline
    MV3D~\cite{Chen2016Multi}  & 86.55 & 78.10 &76.67 & 71.29 & 62.68 & 56.56\\  
    F-PointNet~\cite{qiFrustumPointNets3D2018}  & 88.16 & 84.02 & 76.44 & 83.76 & 70.92 & 63.65\\  
    VoxelNet~\cite{zhouVoxelNetEndtoEndLearning2018}  & 89.60 & 84.81 & 78.57 & 81.97 & 65.46 & 62.85\\
    SECOND~\cite{yan2018second} & 89.96 & 87.07 &79.66 & {\bf 87.43} & 76.48 & 69.10\\
    PointPillars~\cite{langPointPillarsFastEncoders2019} & N/A & 87.7 & N/A & N/A & 77.4 & N/A\\
    \hline 
    Ours & {\bf 93.29} & {\bf 88.90} & {\bf 87.72} & 87.21 & {\bf 77.58} & {\bf 71.79}\\ 
    % & 96.50 & 91.66 & 89.13
    % 0.932945 0.888977 0.87717
    \hline
    \end{tabular}
    \caption{Performance (AP) of BEV and 3D on KITTI validation set for \textit{Car}. Our method achieves the state of the art in most case.}
    \label{tab:car_valid}
    \vspace{-5pt}
    \end{table}
    
    \begin{table}
    % \vspace{-5pt}
    \small
    \centering  
    \begin{tabular}{C{2cm}|C{1cm}|C{1cm} C{1cm} C{1cm}}
    \hline
    {\bf Classes} & {\bf Tasks} & {\bf \textit{Easy}} & {\bf \textit{Moder.}} & {\bf \textit{Hard}}\\
    \hline
    { \textit{Car}} & 2D & 96.50 & 92.04 & 89.34 \\
    \hline
    \multirow{3}*{ \textit{Pedestrian}} & 2D & 76.53 & 70.75 & 68.37 \\
    % 0.756487 0.697598 0.6525
    & BEV & 75.65 & 69.76 & 65.25  \\
    & 3D & 69.13 & 64.81 & 59.42 \\
    \hline
    \multirow{3}*{ \textit{Cyclist}} & 2D & 93.90 & 80.54 & 77.41 \\
    % 0.899786 0.771259 0.672713
    & BEV & 89.98 & 77.13 & 72.27\\
    & 3D & 87.21 & 73.75 & 68.98 \\
    \hline
    \end{tabular}
    \caption{Performance on KITTI validation set for \textit{Pedestrian}, \textit{Cyclist} in 2D, BEV and 3D tasks; and \textit{Car} in 2D task.}
    \label{tab:other_valid}
    \vspace{-11pt}
    \end{table}
    
    \textbf{Quantitative Analysis.}
    We compare with 3D object detectors in three types: a) LiDAR \& image based approaches; b) LiDAR only two-stage approaches and c) one-stage approaches, shown in three columns respectively in Tab.~\ref{tab:test_results}. 
    Most methods in a) and b) are relatively slow for inference.
    From the table we see that HVNet outperforms all other approaches in \textit{mAP} and \textit{Cyclists}. 
    HVNet also achieves attractive performance for \textit{Car} and \textit{Pedestrian} under a real-time runtime, even when compared with two-stage approaches. 
    Among one-stage approaches, HVNet achieves the state of the art  in \textit{Car} and \textit{Cyclist}, leading the second best HRI-VoxelFPN~\cite{wangVoxelFPNMultiscaleVoxel2019} and PointPillars~\cite{langPointPillarsFastEncoders2019} by over 1.61\% and 8.44\% respectively in \textit{moderate}. More details for our test results are in KITTI \href{http://www.cvlibs.net/datasets/kitti/eval_object.php?obj_benchmark=bev}{leaderboard}. Furthermore, we draw the performance vs. speed figure in Fig.~\ref{fig:P-R} according KITTI leaderboard to have more intuitive comparison.
    
\begin{figure}
    % \vspace{-5pt}
    \centering
    \includegraphics[width=7.8cm]{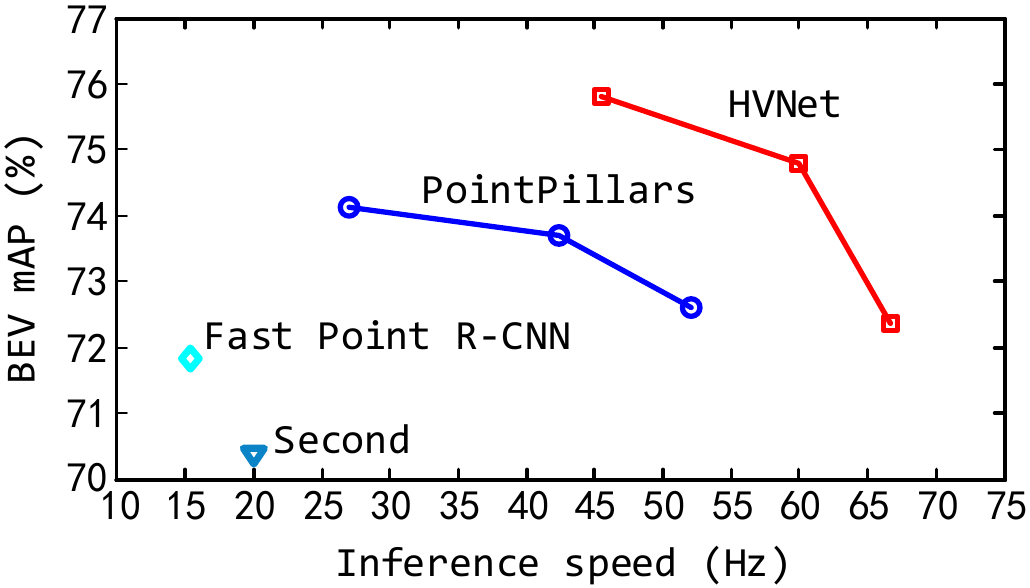}
    \caption{Voxel scale study on BEV of KITTI validation set. For PointPillar we use our own implementation in Pytorch. Blue circle shows the results of PointPillar. We choose voxel scale at \{$0.16^2$, $0.20^2$, $0.24^2$\}$m^2$. Red rectangle illustrates results from Hybrid Voxel Feature Extractor. From left to right, we use voxel scales at \{0.1, 0.2\}$m$ ,\{0.2, 0.3\}$m$, \{0.2, 0.4\}$m$, feature projection scales at 0.2$m$, 0.3$m$, 0.4$m$ respectively }
    \label{fig:voxel_scale}
    \vspace{-10pt}
    \end{figure}

    Only a few methods expose the results on the validation set. 
    The comparison results for \textit{Car} are reported in Tab.~\ref{tab:car_valid}. Among methods that expose results, our approach achieves the best performance in both BEV and 3D tasks.
    As almost no currently published method presents validation results for \textit{Pedestrian} and \textit{Cyclist}, we show only our validation results for these two classes in Tab~\ref{tab:other_valid} in all the three tasks: 2D, BEV and 3D.
    Overall, evaluation both on test and validation set shows that our approach can produce high-accuracy detections with a fast inference speed.
    Besides that, we conduct experiments on our Hybrid Voxel Feature Extractor architecture at different settings, compared with PointPillar through different grid size in Fig.~\ref{fig:voxel_scale}. We can easily see the power of our architecture. Compared with best result from PointPillar at grid size 0.16m, our model show great advantages at both mAP and inference speed. Even at a coarse scale for feature projection scale of grid size 0.4$m$, our model can achieve comparable results of PointPillar at grid size 0.24$m$ and save a lot of runtime cost.
    
    \begin{table*}
    \centering
    \small
    \vspace{-13pt}
    \begin{tabular}{l|c c c|c c c|c|c|c|c|c c c}
    \hline
    \multirow{2}*{\bf Model} & \multicolumn{3}{c|}{\bf voxel scale $S_T$} & \multicolumn{3}{c|}{\bf projection scale $S_R$} & \multirow{2}*{\bf AVFE} & \multirow{2}*{\bf FPN} & \multirow{2}*{\bf FFPN} & \multirow{2}*{\bf $\alpha$} & \multicolumn{3}{c}{\bf BEV mAP (\%)} \\
    \cline{2-7}
    \cline{12-14}
        & $0.5S$ & $1S$ & $2S$ & $1S$ & $2S$ & $4S$ & & & & & { \textit{Easy}} & { \textit{Moder.}} & { \textit{Hard}}  \\
    \hline
    Baseline & \checkmark & & & \checkmark & &  & & & & 0.25 & 79.39 & 71.7 & 66.77 \\
    + Attention & \checkmark & & & \checkmark & &  & \checkmark & & & 0.25 & 81.36 & 73.76 & 69.18  \\
    + FPN & \checkmark & & & \checkmark & &  & \checkmark & \checkmark & & 0.25 & 81.84 & 74.34 & 69.83 \\
    + FFPN & \checkmark & & & \checkmark & &  & \checkmark &  & \checkmark & 0.25 & 82.39 & 75.18 & 70.78  \\
    * Focal loss & \checkmark & & & \checkmark & &  & \checkmark &  & \checkmark & 0.75 & 82.82 & 75.71 & 71.54   \\
    + Scale 2* & \checkmark & \checkmark & & \checkmark & &  & \checkmark &  &\checkmark  & 0.75 & 84.34 & 76.88 & 73.01   \\
    + Scale 2 & \checkmark & \checkmark & & \checkmark & \checkmark &  & \checkmark &  & \checkmark & 0.75 & 85.26 & 77.88 & 74.46   \\
    + Scale 3 & \checkmark & \checkmark & \checkmark & \checkmark & \checkmark & \checkmark & \checkmark &  & \checkmark & 0.75 & \bf 86.31 & \bf 78.59 & \bf 75.08  \\
    \hline
    \end{tabular}
    \caption{Ablation study on KITTI validation set for the \textit{mAP}. $0.5S$ means the scale of $0.5$. $\Delta$ mAP denotes the change in \textit{Moderate} mAP compared with the corresponding controlled experiment. 
    The maximum improvement is achieved by increasing $S_T$ and $S_R$.
    }
    \label{tab:ablation}
    \vspace{-10pt}
    \end{table*}

    % \begin{figure}
    % \vspace{-5pt}
    % \centering
    % \includegraphics[width=8.3cm]{failure}
    % \caption{Failure cases on KITTI validation set.}
    % \label{fig:failure}
    % \vspace{-15pt}
    % \end{figure}
    
    \textbf{Qualitative Analysis.}
    We present several 3D detections on the KITTI validation set along with projected 2D bounding boxes on 2D image in Fig.~\ref{fig:qualitative}. 
    HVNet can produce high-quality 3D detections for all classes via a single pass of the network.
    Moreover, good detections are also given by HVNet for scenes that have point cloud occlusion or strongly dense objects
    % , such as the left bottom example in Fig.~\ref{fig:qualitative}
    .
    Generally, these qualitative results demonstrate the effectiveness of our approach.
    
    % There are two main types of failure cases: a) the missing cases in the ground truth of KITTI dataset, such as the missing pedestrian and cyclist in the penult example of Fig.~\ref{fig:qualitative}.
    % HVNet have surprisingly some detections in these cases.
    % b) some actual failure cases visualized in Fig.~\ref{fig:failure}.
    % In the first example, a pickup truck is detected into two separates cars, caused by the lack of points in the middle of the car. The waste bin and grass in the second example are regarded as pedestrian and cyclist respectively, both of which are very confusing when using LiDAR only.

    \vspace{-3pt}
    \subsection{Ablation Studies}
    \label{sec:ablation}
    \vspace{-5pt}
    
    \textbf{Multiple components study.}
    To analyze how different components contribute to the final performance, we conduct an ablation study on KITTI validation set. 
    We use the BEV mean AP of three categories (\textit{Car}, \textit{Pedestrian} and \textit{Cyclist}) as the evaluation matrix, shown in Tab.~\ref{tab:ablation}.
    Our baseline model, shown in the first line of Tab.~\ref{tab:ablation}, is a detector with a single scale feature extractor and a SSD~\cite{liu2016ssd} like backbone. 
    Besides, the VFE layers~\cite{zhouVoxelNetEndtoEndLearning2018} replace the AVFE and AVFEO layers. 
    It's given that adding attention to VFE layer is able to bring a 2.06 mAP gain in BEV \textit{Moderate} compared with baseline. 
    FPN is effective with a 0.58 gain but the proposed FFPN bring a larger improvement of a 1.42 gain. 
    We also adopt focal loss for classification. However, the default $\alpha$ results in the confidence score degradation for \textit{Pedestrian} and \textit{Cyclist}. 
    Therefore, we change $\alpha=0.75$ by experiments. As to the Hybrid Voxel Feature Extractor, increasing voxel/feature projection scale number to 2 gives the maximum performance boost of \textbf{2.17} in mAP. 
    Furthermore, keeping going up to 3 scales gives another 0.71 gain.

    \begin{figure}
    \vspace{-5pt}
    \centering
    \includegraphics[width=7.5cm]{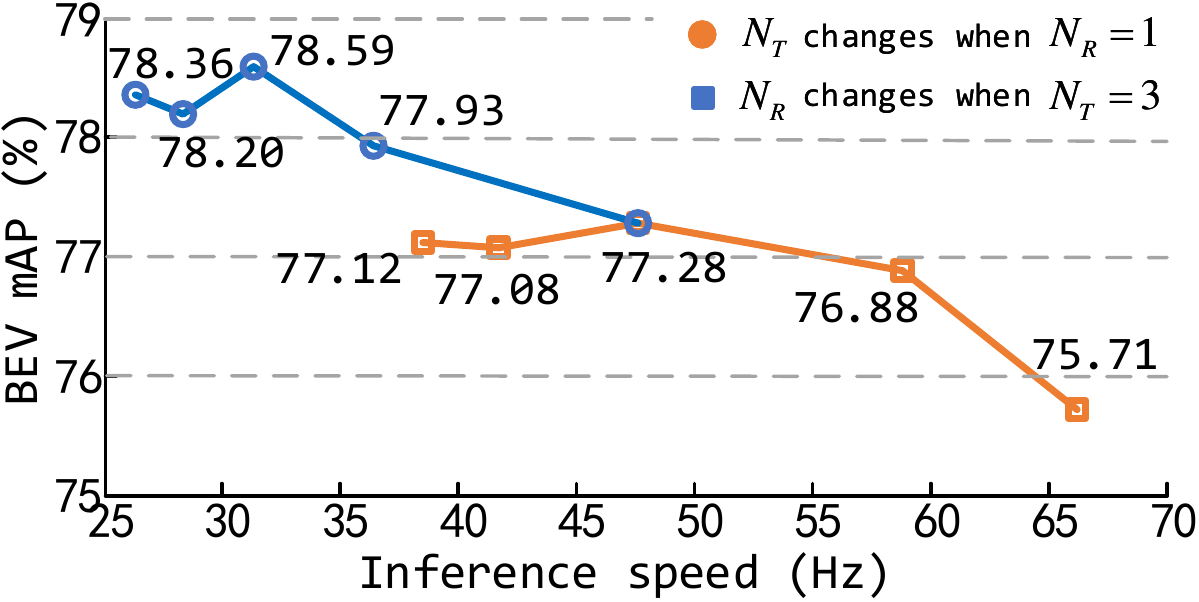}
    \caption{Ablation study on KITTI validation set for hybrid scale. The jacinth line means $N_T$ changes from 5 to 1 when $N_R=1$. The blue line means $N_R$ changes from 5 to 1 when $N_T=3$. 
    % The points from left to right are represent model results with scale numbers from 5 to 1.
    }
    \label{fig:ablation}
    \vspace{-5pt}
    \end{figure}
    
    \textbf{Hybrid Voxel Feature Extractor.}
    % The proposed Hybrid Voxel Feature Extractor is able to zoom multiple scales in and out. 
    Given that the Hybrid Voxel strategy plays an important role, it is important to make $S_T$ and $S_R$ enough for feature encoding while not consuming much computation. 
    Thus, we conduct a series of experiments with various scale number. 
    Note that as the scale number increases, the block number in backbone in Fig.~\ref{fig:backbone} increases as well. 
    Shown in Fig.~\ref{fig:ablation}, it's a nice trade-off between speed and performance when ${N_T} = {N_R} = 3$, demonstrated that the effectiveness of scale projection between $S_T$ and $S_R$. 
    Furthermore, we visualize the ${\bf{H}}^{S_T}$ with/without attention in Fig.~\ref{fig:features}, which shows that the activation of target region is greater with attention strategy.
    
    \begin{table}[!h]
    \vspace{-3pt}
    \centering
    \small
    \begin{tabular}{c |c |c | c}
    \hline
    VFE~\cite{langPointPillarsFastEncoders2019} & HSV & Multi-class Head & Time (ms) \\
    \hline
    \checkmark & $\times$& $\times$& 34\\
    $\times$& \checkmark & $\times$ & 32\\
    $\times$& \checkmark & \checkmark & 35\\
    \hline
    \end{tabular}
    \caption{Inference speed for HVNet with different module.}
    \label{tab:speed}
    \vspace{-10pt}
    \end{table}
    
    \begin{figure}
    \vspace{-3pt}
    \centering
    \includegraphics[width=7.8cm]{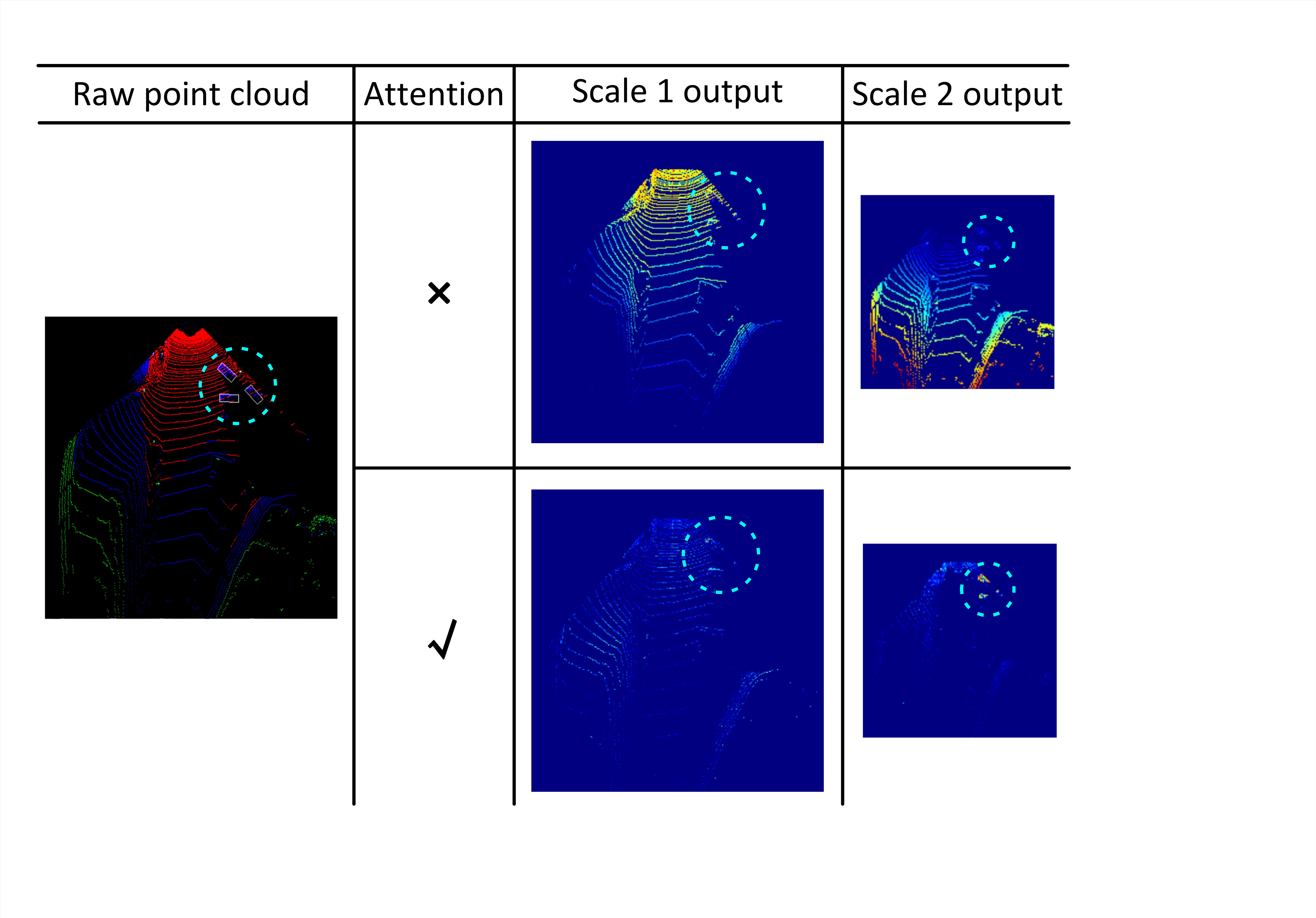}
    \caption{Multi-scale features ${\bf{H}}^{S_T}$ with/without attention. With attention mechanism, our output feature map can suppress the background area and enhance the shape feature of objects}
    \label{fig:features}
    \vspace{-10pt}
    \end{figure}
    
    \textbf{Inference speed.}
    The inference time of HVNet is 32\textit{ms} for a single class in average on a 2080Ti GPU where the Hybrid Voxel Feature Extractor takes 12\textit{ms}, the backbone network takes 11\textit{ms} and the head with NMS takes the left 9\textit{ms}. The required time for each module is changed with the number of input points. 
    In our approaches, HSV and index-based implementation are proposed to accelerate the feature encoding, whose effectiveness is shown in Tab.~\ref{tab:speed}. We employ the VFE layer in PointPillars \cite{langPointPillarsFastEncoders2019} as baseline. 
    Utilize the HSV and index-based implementation save 2ms in average.
    Furthermore, the head in our model only takes extra 3ms for extending multi-class detection in one forward.
    % is for multi-class in one forward, which is able to save 3ms for the computation of co-category part.

    % \subsection{Visualization}
    % \label{sec:visualization}
    \vspace{-3pt}
    \section{Conclusion}
    \vspace{-5pt}
    In this work, we propose HVNet, a novel one-stage 3D object detector. 
    HVNet aggregates hybrid scale voxel grids into unified point-wise features, and then projects them into different scale pseudo-image features under the guidance of attention knowledge.
    The key to HVNet is that it decouples the feature extraction scales and the pseudo-image projection scales.
    Further more, a backbone with feature fusion pyramid network takes pseudo-images and fuses features to generate compact representations for different categories.
    % Finally, we utilize a parallel head to detect all categories within one forward.
    Experimental studies show that our method achieves state-of-the-art \textit{mAP} with a real-time speed.

{\small
\bibliographystyle{ieee_fullname}
\bibliography{egbib}
}

\clearpage
\setcounter{section}{0}
\setcounter{figure}{0}
\setcounter{table}{0}

\begin{center}
    \LARGE{\textit{Supplementary Material}}
\end{center}
% \large{\textit{Supplementary Material}}
% \\\large{\textbf{HVNet: Hybrid Voxel Network for LiDAR Based 3D Object Detection}}

\vspace{20pt}

% \begin{document}

% %%%%%%%%% TITLE
% \title{Supplementary Material for\\\textbf{HVNet: Hybrid Voxel Network for LiDAR Based 3D Object Detection}}

% \author{Anonymous CVPR submission}
% \date{Paper ID 923}

% \maketitle
%\thispagestyle{empty}
% \vspace{-20pt}
% \section{Contents}
% In this supplementary material, we provide:
% \begin{itemize}
% \small
% %  \setlength{\itemsep}{4pt}
% %  \setlength{\parskip}{0pt}
% %  \setlength{\parsep}{0pt}
% %  \item Detailed performance analysis in Sec.~\ref{sec:analyzes}.
% %  \item Other Attempts of HVNet in Sec.~\ref{sec:attempts}.
%  \item More About the Attention Mechanism in Sec.~\ref{sec:attention}.
%  \item More Ablation Studys in Sec.~\ref{sec:ablation}.
%  \item Quantitative results of our approach in Sec.~\ref{sec:quantitative}.
%  \item Qualitative results of our approach in Sec.~\ref{sec:analyzes}.
%  \item Feature map visualization for different scale $S_R$.
% \end{itemize}

% \vspace{-20pt}  

% \section{Other Attempts}
% \label{sec:attempts}

% \textbf{IoU branch}

\begin{table*}
\centering
\begin{tabular}{c  |c  c  c | c  c  c | c  c  c}
\hline
\multirow{2}*{\bf Method} & \multicolumn{3}{c|}{\bf \textit{Car} BEV AP (\%)} & \multicolumn{3}{c|}{\bf \textit{Pedestrian} BEV AP (\%)} & \multicolumn{3}{c}{\bf \textit{Cyclist} BEV AP (\%)} \\ \cline{2-10}
        & {\textit{Easy}} & {\textit{Moder.}} & {\textit{Hard}} & {\textit{Easy}} & {\textit{Moder.}} & {\textit{Hard}} & {\textit{Easy}} & {\textit{Moder.}} & {\textit{Hard}} \\  
    \hline

Corner Loss & 93.29   &    88.90     &  87.72   & 75.65
    & 69.76 & 65.25   & 89.98 & 77.13 & 72.27 \\
Pose Loss & 93.07   & 87.67   & 84.76   & 76.13   & 68.37   & 63.77   & 87.26   & 73.34   & 69.72 \\
\hline
\end{tabular}
\caption{Performance of bird eye views on KITTI val set.} %\protect\footnotemark[1]
\label{tab:loss_test}
\end{table*}

\section{More About the Attention Mechanism}
\label{sec:attention}
The attention layer can be thought of a kernel method for the input feature: 
$X'{\rm{ = }}WG \cdot W'X$
% \begin{equation}
%     X'{\rm{ = }}WG \cdot W'X
% \end{equation}
, where $G$ represents attention feature which is described in Equ. \textbf{2} in the paper, $W$ and $W'$ are linear weights. Since that $G$ contains a part of $X$ in the first layer, the attention mechanism can be regarded as a second-order kernel for input $X$.
In experiment Fig. \textbf{10} we show that with attention the output feature map suppresses the background and enhances the shape feature of object. 

\section{Ablation Studys}
\label{sec:ablation}

\subsection{Corner Loss}
Most of existing methods often regress $x$, $y$, $z$, $l$, $w$, $h$, $theta$ of the 3D bounding box of the object, which we name pose loss. 
We first analyze the difference between corner loss and pose loss. For simplicity, we use our best model on KITTI validation. 
We can find that our corner loss perform better than original loss from Tab. ~\ref{tab:loss_test}. 
According to our experiment, learning $theta$ directly is a little difficult for pure point cloud data. Also, there is a discontinuity for angle 0 and $2\times \pi$. We could add some normalization to alleviate this problem. However, there are still some corner cases such as the gap between 0 and $2 \times \pi - \epsilon$, where $\epsilon$ being a small number.

\subsection{Loss Varietas}
We also tried the IoU branch and the IoU/GIoU Loss in our HVNet. For IoU branch, we add a parallel convolution layer in the detection head to predict the IoU of each anchor, and then multiply the IoU prediction and cls score in the inference time. This strategy performs well in valid set but fails in test set. The IoU Loss is use the IoU as loss directly, which fails in both val and test set.

Besides, we also tried the uncertain loss to learn the weight of different item of loss, such as corner loss, elevation loss and cls loss. However, it does not improve the result work as well.

% \vspace{-15pt}
\section{Quantitative Analysis}
\label{sec:quantitative}

We have provide all results on 3D/BEV/3D on the KITTI validation set, but only BEV results on the test set due to the space limit. The overall quantitative results on test are added in Tab. \ref{tab:test_supp}.

\section{Qualitative Analysis}
\label{sec:analyzes}

In this section, we provide more qualitative visual results of our method on KITTI validation/test set.

\subsection{Qualitative Results}
We provide more visual results of KITTI validation set in Fig~\ref{fig:qualitative}. 
Besides, the results of KITTI test set is shown in Fig~\ref{fig:qualitative_test}.
Generally, these qualitative results demonstrate the effectiveness of our approach.

\subsection{Failure Case.}
\begin{figure*}
\centering
\includegraphics[width=1.0\textwidth]{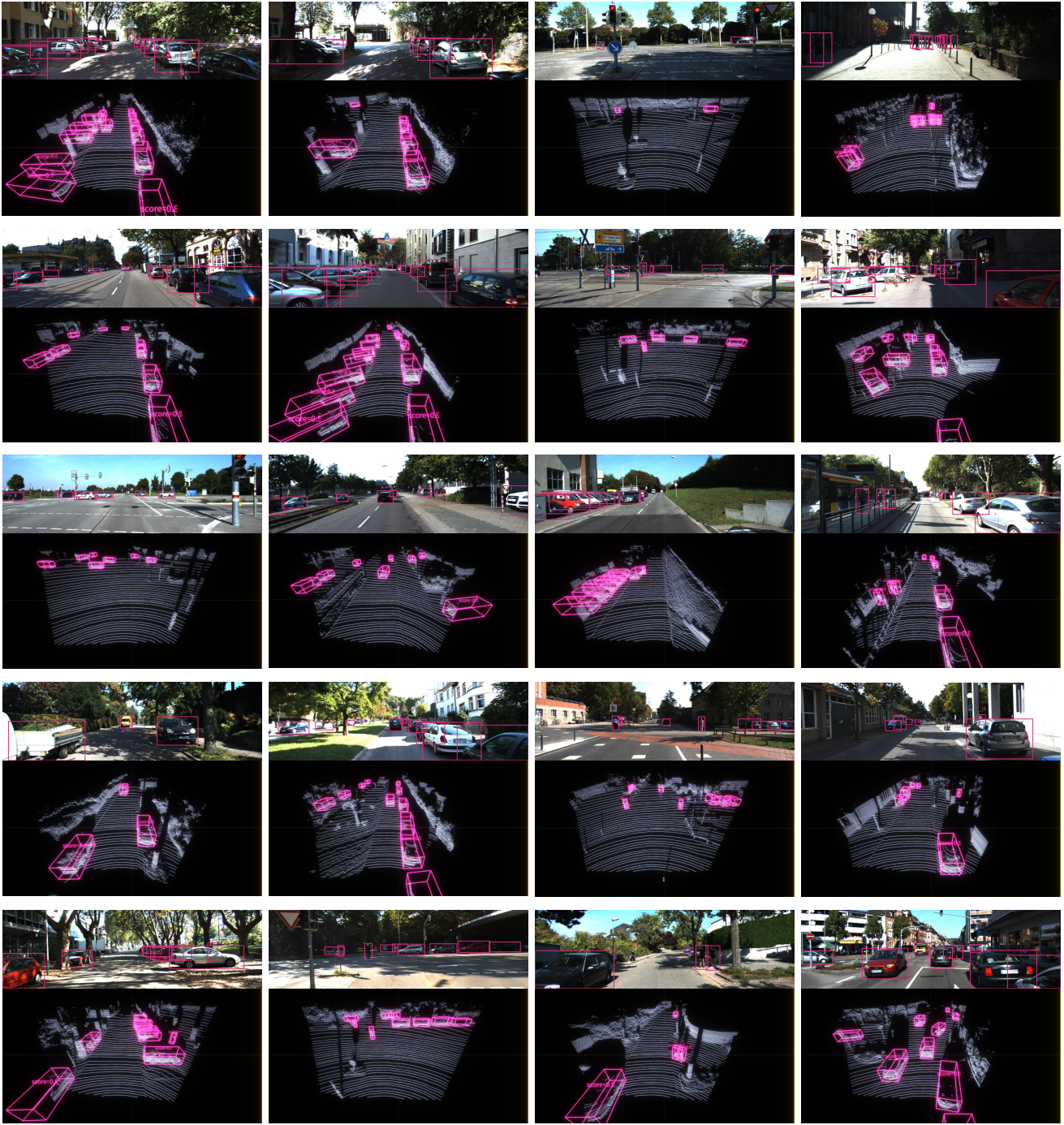}
\caption{Qualitative analysis on KITTI test set with Kitti Viewer. We show 3D bounding boxes on point cloud along with projected 2D bounding boxes on image. In each image, red boxes indicates detections by HVNet.}
% Qualitative analysis on KITTI. We show a bird’s eye view of the LIDAR point cloud (top), as well as the 3D bounding boxes projected into the image for clearer visualization. Note that our method only uses LIDAR. We show ground truth (gray) and predicted boxes for car (orange), cyclist (red) and pedestrian (blue). The box orientation is shown by a line from the bottom center to the front of the box.
\label{fig:qualitative_test}
\end{figure*}

\begin{figure*}
% \vspace{-5pt}
\centering
\includegraphics[width=1.0\textwidth]{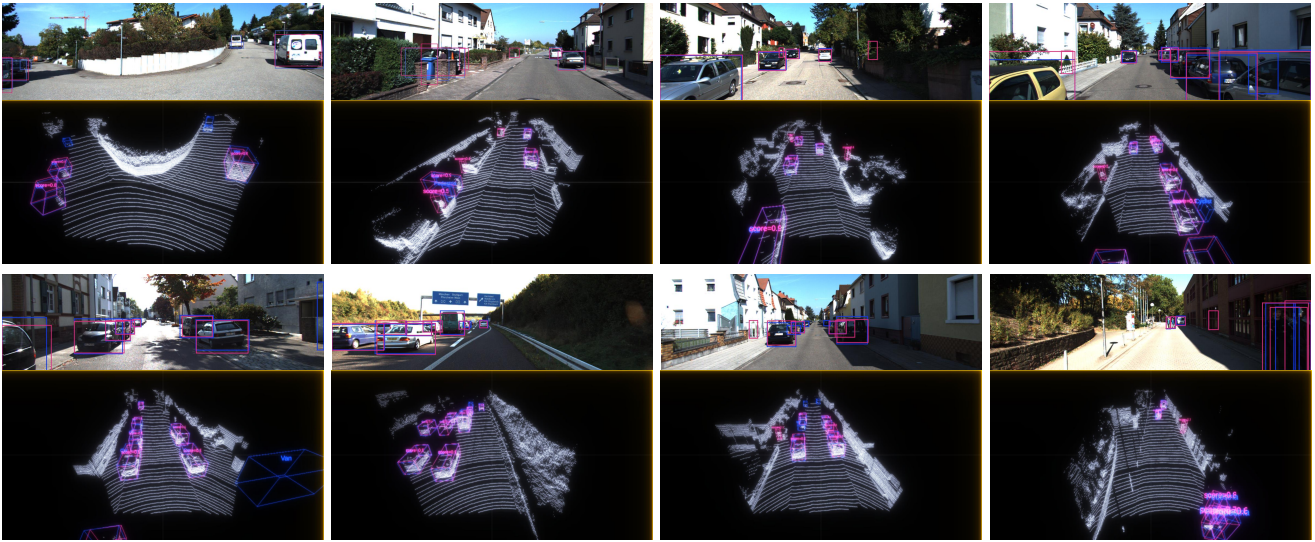}
\caption{Failure cases on KITTI validation set.}
\label{fig:failure}
% \vspace{-15pt}
\end{figure*}

There are two main types of failure cases: 
\begin{itemize}
    \item The missing cases in the ground truth of KITTI dataset, such as the missing pedestrian and cyclist in the second example of Fig.~\ref{fig:qualitative} and~\ref{fig:qualitative_test}.
    HVNet have surprisingly detects some of the objects in these cases.
    \item Some actual failure cases of HVNet, visualized in Fig.~\ref{fig:failure}. In the first example, a pickup truck is detected into two separates cars, caused by the lack of points in the middle of the car. The waste bin and grass in the second example are regarded as pedestrian and cyclist respectively, both of which are very confusing when using LIDAR only.
\end{itemize}

\subsection{Feature Learning of $S_R$}
We visualize the features ${\bf{I}}^{S_R}$ of scale $S_R$ in Fig.~\ref{fig:features}. Each point cloud has features of three scales as a counterpart. We choose the first channel of each feature and then draw it up.

\begin{figure*}
\centering
\includegraphics[width=1.0\textwidth]{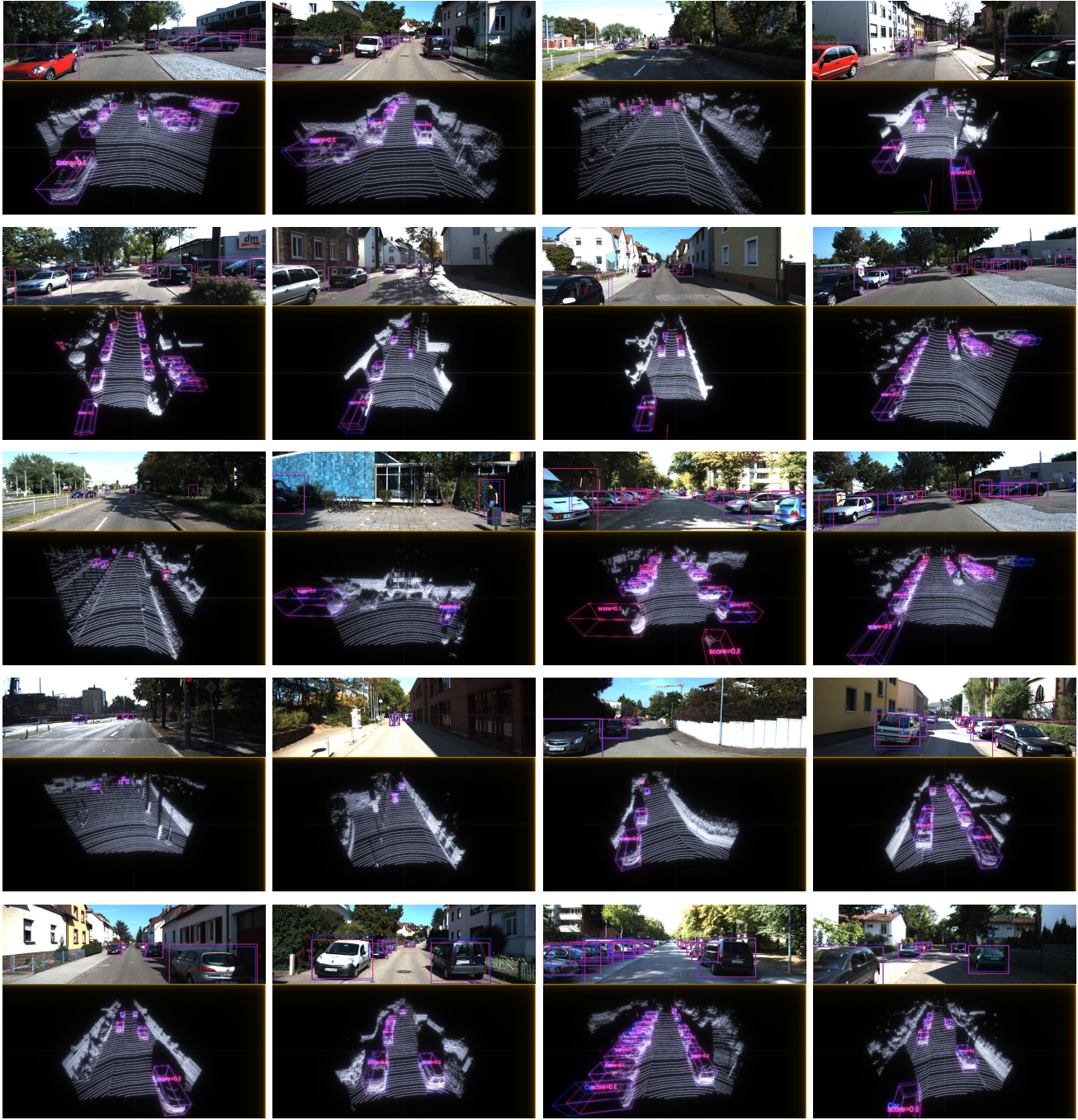}
\caption{Qualitative analysis on KITTI validation set with Kitti Viewer~\cite{yan2018second}. We show 3D bounding boxes on point cloud along with projected 2D bounding boxes on image. In each image, blue boxes indicates the ground truth, red boxes indicates detections by HVNet.}
% Qualitative analysis on KITTI. We show a bird’s eye view of the LIDAR point cloud (top), as well as the 3D bounding boxes projected into the image for clearer visualization. Note that our method only uses LIDAR. We show ground truth (gray) and predicted boxes for car (orange), cyclist (red) and pedestrian (blue). The box orientation is shown by a line from the bottom center to the front of the box.
\label{fig:qualitative}
\end{figure*}

\begin{table*}
\centering
\begin{tabular}{c  |c  c  c | c  c  c | c  c  c}
\hline
\multirow{3}*{\bf Method} & \multicolumn{3}{c|}{\bf \textit{Detection} AP (\%)} & \multicolumn{3}{c|}{\bf \textit{Orientation} AP (\%)} & \multicolumn{3}{c}{\bf \textit{3D Detection} AP (\%)} \\ \cline{2-10}
        & {\textit{Easy}} & {\textit{Moder.}} & {\textit{Hard}} & {\textit{Easy}} & {\textit{Moder.}} & {\textit{Hard}} & {\textit{Easy}} & {\textit{Moder.}} & {\textit{Hard}} \\  
    \hline

Car  & 96.02   &  92.11  &  86.77  & 42.61 & 42.98 & 39.68   & 84.49 & 72.45  & 66.66 \\
Pedestrian &  64.37   & 55.56   & 52.98   & 35.07   & 29.17  & 27.57  & 47.98   & 41.44   & 38.57 \\
Cyclist & 86.13    & 76.93   & 70.07   & 44.10   & 38.92  & 34.81  & 80.78  & 66.45  & 58.76 \\
\hline
\end{tabular}
\caption{Overall quantitative results on KITTI test set.} %\protect\footnotemark[1]
\label{tab:test_supp}
\end{table*}

\begin{figure*}
% \vspace{-5pt}
\centering
\includegraphics[width=1.0\textwidth]{features_supp}
\caption{Multi-scale features ${\bf{I}}^{S_R}$. Three output features is set in our experiments.}
\label{fig:features}
% \vspace{-15pt}
\end{figure*}

\end{document}